\documentclass[journal]{IEEEtran}

\usepackage[utf8]{inputenc}
\usepackage{color}
\usepackage{xcolor}
\usepackage{array}
\usepackage{verbatim}
\usepackage{float}
\usepackage{amsmath}
\usepackage{amsthm}
\usepackage{amssymb}
\usepackage{graphicx}
\usepackage{longtable}
\usepackage{multirow}
\usepackage{booktabs}
\usepackage[unicode=true,
bookmarks=false,
breaklinks=false,pdfborder={0 0 1},colorlinks=false]
{hyperref}
\hypersetup{
	colorlinks,bookmarksopen,bookmarksnumbered,citecolor=blue,urlcolor=blue}
\usepackage{cite}

\usepackage{lipsum}
\usepackage{mathtools}
\usepackage{cuted}

\usepackage{algorithmic}
\usepackage{longtable}

\floatstyle{ruled}
\newfloat{algorithm}{tbp}{loa}
\providecommand{\algorithmname}{Algorithm}
\floatname{algorithm}{\protect\algorithmname}

\makeatletter
\let\oldforeign@language\foreign@language
\DeclareRobustCommand{\foreign@language}[1]{%
	\lowercase{\oldforeign@language{#1}}}

\let\oldforeign@language\foreign@language
\DeclareRobustCommand{\foreign@language}[1]{%
	\lowercase{\oldforeign@language{#1}}}

\ifCLASSINFOpdf
\else
\fi

\@ifundefined{showcaptionsetup}{}{%
	\PassOptionsToPackage{caption=false}{subfig}}
\usepackage{subfig}

\usepackage{balance}

\ifCLASSINFOpdf
\else
\fi

\ifCLASSINFOpdf
\else
\fi

\def\ps@IEEEtitlepagestyle{%
	\def\@oddhead{\parbox[t][\height][t]{\textwidth}{\centering \scriptsize
			Personal use of this material is permitted. Permission from the author(s) and/or copyright holder(s), must be obtained for all other uses. Please contact us and provide details if you believe this document breaches copyrights.\\
			\noindent\makebox[\linewidth]{}
		}\hfil\hbox{}}%
	\def\@evenhead{\scriptsize\thepage \hfil \leftmark\mbox{}}%
	\def\@oddfoot{\parbox[t][\height][l]{\textwidth}{
			\vspace{-20pt}{\rule{\textwidth}{0.4pt}}\\ \footnotesize{\bf{\footnotesize\textcolor{red}{M. Khalifa and H. A. Hashim, "Dual Quaternion-Based Unscented Kalman Filter with Visual Inertial Odometry for Navigation in GPS-Denied Environments," Measurement, pp. 121964, 2026.}}} doi: \href{https://doi.org/10.1016/j.measurement.2026.121964}{10.1016/j.measurement.2026.121964}\\\\
			\noindent\makebox[\linewidth]
		}\hfil\hbox{}}%
	\def\@evenfoot{\MYfooter}}

\makeatother
\begin{document}
	\bstctlcite{IEEEexample:BSTcontrol}

\title{Dual Quaternion-Based Unscented Kalman Filter with Visual Inertial Odometry for Navigation in GPS-Denied Environments}

\author{Mohamed Khalifa and Hashim A. Hashim
	\thanks{This work was supported in part by the National Sciences and Engineering Research Council of Canada (NSERC), under the grants RGPIN-2022-04937.}
	\thanks{M. Khalifa and H. A. Hashim are with the Department of Mechanical
		and Aerospace Engineering, Carleton University, Ottawa, Ontario, K1S-5B6,
		Canada (e-mail: hhashim@carleton.ca).}
}



\maketitle
\begin{abstract}
Reliable navigation in GPS-denied environments remains a fundamental challenge in robotics, aerospace, and autonomous vehicle applications. This paper presents a Dual Quaternion-Based Unscented Kalman Filter (DQUKF) equipped with a Visual Inertial Odometry (VIO) algorithm for accurate state estimation enabling navigation in GPS denied locations. The proposed framework formulates the DQUKF in an error state manner, where the nominal pose is represented by a unit dual quaternion and the local pose error is represented by a 6-dimensional twistor parameterization used for sigma point generation, covariance propagation, and measurement correction. In parallel, the VIO algorithm tracks features across image frames, synchronizes measurements between the IMU and camera, and provides visual constraints that complement inertial propagation. Simulation results on the EuRoC MAV dataset show that the proposed DQUKF converges under high initialization uncertainty and achieves a position RMSE of 0.2584~m in the difficult flight sequence, outperforming the benchmark filters.
\end{abstract}

\begin{IEEEkeywords}
Unscented Kalman Filter, Visual-Inertial Odometry, dual quaternion,
sensor fusion, GPS-denied navigation 
\end{IEEEkeywords}

\section{Introduction}
Micro aerial vehicles (MAVs) operating in GPS-denied environments
rely on accurate state estimation for navigation and control. Visual-inertial
odometry (VIO), which fuses camera and inertial measurements, has
become a widely adopted solution due to the complementary characteristics
of these sensors \cite{hash2025,Huang2019,hash2023_IJC_UAV_NavCont}. Inertial
measurements provide high-rate motion information but suffer from
drift over time, while visual measurements constrain long-term motion
but are sensitive to environmental conditions. By combining these
sensing modalities, VIO enables reliable estimation of motion in challenging
scenarios \cite{MurArtal2017,hash2022_TSMCS_SLAM}. In tightly coupled
formulations, visual feature observations are incorporated directly
into the estimation process, allowing inertial and visual information
to be fused at the measurement level. This integration improves estimation
accuracy and robustness compared to loosely coupled approaches, particularly
under rapid motion, feature degradation, or visually challenging conditions.
\cite{hashim2025advances, mourikis2007,Chen2018,hashim2026insects,Zeyuan2026_star_sensor,hash2021_ACC_GPSNav}.
Among the various estimation strategies, filter-based methods remain
attractive for real-time VIO due to their recursive structure and
computational efficiency \cite{Usenko2017}. In particular, Kalman
filtering and its nonlinear extensions provide a principled framework
for fusing multi-sensor data while accounting for process and measurement
uncertainty \cite{Julier2004}. However, beyond the choice of filtering
technique, the representation of attitude and rigid-body pose plays
a critical role in estimator performance. The choice of parameterization
directly affects numerical stability, constraint handling, and the
consistency of uncertainty propagation within nonlinear filters. In
addition to the choice of filtering technique, the parameterization
of pose plays a central role in nonlinear estimation. For attitude
estimation, minimal representations such as modified Rodrigues parameters
(MRPs) \cite{Schaub2003} have been widely used due to their ability
to represent rotation using three parameters while avoiding the unit-norm
constraint associated with quaternions. These representations are
typically employed to describe local attitude errors within filtering
frameworks, enabling consistent uncertainty propagation and correction.
Extending this idea to rigid-body motion, dual quaternion representations
can be combined with minimal local parameterizations to represent
pose perturbations in a lower-dimensional space. In particular, dual
extensions of modified Rodrigues parameters \cite{Filipe2014} provide
a compact six-dimensional representation of local pose error, which
can be interpreted as a twistor representation of rigid-body motion.
Such formulations offer a promising approach for combining the geometric
consistency of dual quaternions with the numerical advantages of minimal
local parameterizations.

\subsection{Related Work}

Modern filter-based state estimation is built on the concepts of Kalman
filtering. The linear Kalman filter provided an optimal estimation
for a linear system with noisy measurements \cite{Kalman1960}. From
there, improvements where made to handle nonlinear systems more efficiently,
leading to the development of the EKF and UKF \cite{Gruber1967}\cite{Julier1997}.
Alongside advances in filtering techniques, prior work has also investigated
different parameterizations for representing attitude and rigid-body
motion within estimators. Euler angle based attitude estimation for
spacecraft using Kalman filtering was performed \cite{Farrell1970}.
The multiplicative EKF \cite{Lefferts1982} performs attitude estimation
using unit quaternions while ensuring the measurement update satisfies
the quaternion unit constraint. In \cite{CrassidisMarkley2003}, a
UKF formulation for attitude estimation is presented in which the
global orientation is parameterized using quaternions. Attitude parametrization
and estimation using nonlinear filtering was further surveyed in \cite{Shuster1993}\cite{Crassidis2007}.The
representation of different attitude errors inside the Kalman filtering
framework was explored by \cite{Markley2003}. In the literature,
unit quaternions are the most commonly used parametrization for attitude.
The traditional approach to pose estimation in navigation filters
represents attitude and position as separate states, typically parameterizing
orientation with quaternions and position in Euclidean space. Dual
quaternions provide a compact and geometrically consistent representation
of rigid-body transformations, allowing rotation and translation to
be represented within a single algebraic structure. A dual quaternion
based linear Kalman filter has been employed to estimate a $\mathbb{SE}(3)$
with linear update model with state dependent measurement noise \cite{Rangaprasad2016}.
For nonlinear filtering, dual quaternions have been used with the
iterated extended Kalman filter (IEKF) in \cite{Goddard1998} to estimate
a $\mathbb{SE}(3)$ element with nonlinear update model. Other EKF
formulations using unit dual quaternions were proposed in \cite{BayroCorrochano2000}
and \cite{Zu2014}. In both methods, the measurement update is applied
additively, which does not inherently preserve the unit dual quaternion
constraint and therefore requires normalization. The approach in \cite{Filipe2015}
extends the multiplicative EKF for unit quaternion based attitude
estimation in \cite{Lefferts1982} to pose estimation using unit dual
quaternions while preserving the normalization constraints. In \cite{Sveier2021},
a particle filter for pose estimation based on unit dual quaternions
was proposed. The method employs unit dual quaternions for global
pose representation and dual MRPs for local pose parameterization,
allowing the mean and covariance to be computed without violating
the unit dual quaternion constraints. A dual quaternion UKF for pose
estimation was proposed in \cite{Deng2016}, which is closely related
to the present work. However, the approach assumes that the measurements
are direct outputs of navigation sensors rather than feature based
observations. In visual inertial odometry, feature observations obtained
from images play a crucial role in correcting the drift introduced
by IMU integration, thus improving estimation accuracy and robustness.

Dual quaternions provide a compact and geometrically consistent representation
of rigid-body transformations, encoding six degrees of freedom using
eight parameters. This makes them particularly attractive for robotics
and pose estimation applications. However, as constrained representations,
they require careful handling of uncertainty within filtering frameworks.
Motivated by these considerations, this work employs a unit dual quaternion
representation for the nominal pose, together with a minimal local
pose error parameterization, within a UKF framework to enable consistent
pose estimation for visual inertial navigation.

\subsection{\textcolor{black}{Contribution}}

\textcolor{black}{To address the above-mentioned limitations, this
	work proposes a tightly coupled VIO framework based on a Dual Quaternion
	Unscented Kalman Filter (DQUKF) for state estimation in GPS-denied
	environments. The main contributions of this work are summarized as
	follows: }
\begin{enumerate}
	\item[B1.] \textcolor{black}{A novel DQUKF is proposed for visual-inertial state
		estimation, enabling consistent uncertainty propagation on the pose
		manifold by representing rigid-body motion using unit dual quaternions,
		thereby preserving the geometric structure associated with rigid-body
		transformations and avoiding singularities associated with minimal
		parameterizations. }
	\item[B2.] \textcolor{black}{A twistor-based local pose parameterization is formulated
		through dual MRP to describe rigid-body motion as a six-dimensional
		perturbation of the nominal unit dual quaternion pose. }
	\item[B3.] \textcolor{black}{A tightly coupled visual-inertial fusion framework
		is developed in which raw visual feature observations are directly
		integrated with inertial measurements, effectively reducing IMU drift
		while maintaining computational efficiency. }
\end{enumerate}

\subsection{Structure}

The remainder of the paper is organized as follows: Section \textsc{\eqref{sec:Preliminaries}}
mentions the preliminary concepts and mathematical foundation. Section
\textsc{\eqref{sec:Dual-Quaternion-based-UKF}} discusses the proposed
dual quaternion based UKF. Section \textsc{\eqref{sec:VIO-framework}}
highlights the VIO framework. Section \textsc{\eqref{sec:Results}}
presents the simulation results and Section \textsc{\eqref{sec:Conclusion}}
summarizes the paper.

\section{Preliminaries\label{sec:Preliminaries}}

The set of real numbers with dimensional space $n$-by-$m$ is given
by $\mathbb{R}^{n\times m}$. $I_{n}\in\mathbb{R}^{n\times n}$ and
$0_{n\times n}$ represent the identity matrix and zero matrix with
dimension $n$-by-$n$, respectively. $\vec{0}=[0,0,0,0]^{\top}\in\mathbb{R}^{4}$
is a zero vector. The Euclidean norm of a given column vector $v\in\mathbb{R}^{n}$
is defined as $||v||=\sqrt{v^{\top}v}$. The vehicle world-frame and
body-frame are denoted by $\{\mathcal{W}\}$ and $\{\mathcal{B}\}$,
respectively. The skew-symmetric operator $[\cdot]_{\times}$ of $v\in\mathbb{R}^{3}$
is defined as 
\begin{equation}
	[v]_{\times}=\begin{bmatrix}0 & -v_{3} & v_{2}\\
		v_{3} & 0 & -v_{1}\\
		-v_{2} & v_{1} & 0
	\end{bmatrix}\in\mathfrak{so}(3),\hspace{5mm}v=\begin{bmatrix}v_{1}\\
		v_{2}\\
		v_{3}
	\end{bmatrix}
\end{equation}
where $\mathfrak{so}(3)$ is the set of $3\times3$ skew-symmetric
matrices representing the Lie algebra of the special orthogonal group
$\mathbb{SO}(3)$. The inverse mapping of skew-symmetric operator
is given by ($\text{vex}:\mathfrak{so}(3)\rightarrow\mathbb{R}^{3}$)
\begin{equation}
	\text{vex}([v]_{\times})=v\in\mathbb{R}^{3}
\end{equation}
The orientation of a rigid body in 3D space is described by a rotation
matrix $R\in\mathbb{SO}(3)$, where $\mathbb{SO}(3)$ is the 3D special
orthogonal group. Elements of this group can be represented on a 3D
smooth manifold with specific group operations. The Lie group is mathematically
represented as \cite{hash2019_TSMCS_SO3_Ito_Strat} 
\begin{equation}
	\mathbb{SO}(3)=\{R\in\mathbb{R}^{3\times3}|det(R)=+1,RR^{\top}=I_{3}\}
\end{equation}
For rigid body motion the Special Euclidean Group $\mathbb{SE}(3)$
is defined as \cite{hash2022_TSMCS_SE3_Stoch_Lewis} 
\begin{equation}
	\mathbb{SE}(3)=\left\{ \left.\begin{bmatrix}R & t\\
		0_{1\times3} & 1
	\end{bmatrix}\right|\hspace{2mm}R\in\mathbb{SO}(3),t\in\mathbb{R}^{3}\right\} 
\end{equation}
A dual number $\tilde{a}$ is given by $\tilde{a}=a+\epsilon a'$,
where $a\in\mathbb{R}$ and $a'\in\mathbb{R}$ are real numbers. $\epsilon$
is the dual unit having the following property: $\epsilon\neq0$ and
$\epsilon^{2}=0$ \cite{Veldkamp1976}. The addition and subtraction
of two dual numbers $\tilde{a}_{1}=a_{1}+\epsilon a_{1}'$ and $\tilde{a}_{2}=a_{2}+\epsilon a_{2}'$
is given by $\tilde{a}_{1}\pm\tilde{a}_{2}=a_{1}\pm a_{2}+\epsilon(a_{1}'\pm a_{2}')$.
The multiplication of a dual number with a scalar $\varrho$ is given
by $\tilde{a}\varrho=a\varrho+\epsilon a'\varrho$. The product of
two dual numbers is given by $(a_{1}+\epsilon a_{1}')(a_{2}+\epsilon a_{2}')=a_{1}a_{2}+\epsilon(a_{1}a_{2}'+a_{1}'a_{2})$.
Dual quaternions are obtained by extending quaternions such that their
coefficients are dual numbers instead of real numbers.

\subsection{Quaternions}

The attitude can be parametrized via different means. Euler angle
parametrization is a common approach used to describe the roll, pitch
and yaw of MAVs in pose estimation and navigation problems \cite{Kuipers1999}.
However, it is not singularity-free as it suffers from gimbal lock.
Moreover, it does not provide a unique representation for a given
rotation. Angle-axis and Rodriguez vector parameterization are also
prone to singularities in various configurations \cite{Hashim2019}.
The unit-quaternion is a singularity-free representation of the attitude
with 4 parameters and a unity constraint, satisfying motion in 3D
space. Let $Q$ denote a unit-quaternion vector where $Q=[q_{0},q_{1},q_{2},q_{3}]^{\top}=[q_{0},q_{v}]^{\top}\in\mathbb{S}^{3}$,
with $q_{0}\in\mathbb{R}$ and $q_{v}\in\mathbb{R}^{3}$. The unit
quaternion is defined by 
\begin{equation}
	\mathbb{S}^{3}=\left\{ Q\in\mathbb{R}^{4}\bigm|||Q||=1\right\} 
\end{equation}
The conjugate quaternion is given by \cite{Hashim2019} 
\begin{equation}
	Q^{*}=\begin{bmatrix}q_{0},-q_{1},-q_{2},-q_{3}\end{bmatrix}^{\top}=\begin{bmatrix}q_{0},-q_{v}^{\top}\end{bmatrix}^{\top}\in\mathbb{S}^{3}\label{q_inv}
\end{equation}
The inverse is the same as the conjugate for unit quaternions. The
$\odot$ operator is used for handling quaternion multiplication.
The product of two quaternions $q_{1}=[q_{01},q_{v1}]^{\top}$ and
$q_{2}=[q_{02},q_{v2}]^{\top}$ is given as 
\begin{equation}
	q=q_{1}\odot q_{2}=\begin{bmatrix}q_{01}q_{02}-q_{v1}^{\top}q_{v2}\\
		q_{01}q_{v2}+q_{02}q_{v1}+[q_{v1}]_{\times}q_{v2}
	\end{bmatrix}\in\mathbb{S}^{3}\label{q_multiply}
\end{equation}
The identity quaternion is defined as $q_{I}=[1,0,0,0]^{\top}$ such
that $q\odot q^{-1}=q_{I}$. The rotation matrix can be extracted
from the quaternion via $\mathcal{R}_{q}:\mathbb{S}^{3}\rightarrow\mathbb{SO}(3)$
\cite{Hashim2019}: 
\begin{equation}
	\mathcal{R}_{q}(q)=I_{3}+2q_{0}[q_{v}]_{\times}+2[q_{v}]_{\times}^{2}\in\mathbb{SO}(3)\label{R_q}
\end{equation}

\subsection{Dual Quaternions}

Dual quaternions provide a compact representation of the coupled rotational
and translational motion of a rigid body \cite{Rangaprasad2016}.
A dual quaternion $\tilde{q}$ consists of 8 elements and can be written
as 
\begin{equation}
	\tilde{q}=q\boxplus\epsilon q'=[q_{0},q_{x},q_{y},q_{z},q'_{0},q'_{x},q'_{y},q'_{z}]^{\top}\in\mathbb{R}^{8}\label{dual_quat}
\end{equation}
where $q=[q_{0},q_{x},q_{y},q_{z}]^{\top}\in\mathbb{R}^{4}$ and $q'=[q'_{0},q'_{x},q'_{y},q'_{z}]^{\top}\in\mathbb{R}^{4}$
are quaternions representing the real and dual parts, respectively.
The identity dual quaternion is defined as $\tilde{q}_{I}=q_{I}\boxplus\epsilon\vec{0}\in\mathbb{R}^{8}$.
The conjugate of a dual quaternion is given by $\tilde{q}^{*}=q^{*}\boxplus\epsilon q'^{*}$.
The product of two dual quaternions $\tilde{q}_{1}=q_{1}\boxplus\epsilon q_{1}'$
and $\tilde{q}_{2}=q_{2}\boxplus\epsilon q_{2}'$ is given by 
\begin{equation}
	\tilde{q}_{1}\otimes\tilde{q}_{2}=q_{1}\odot q_{2}\boxplus\epsilon(q_{1}\odot q_{2}'+q_{1}'\odot q_{2})\label{Dq_multiply}
\end{equation}
where $\otimes$ is the dual quaternion multiplication operator. A
dual quaternion represents a rigid body transformation when it satisfies
the unit dual quaternion constraint. A unit dual quaternion satisfies
\[
\tilde{q}\otimes\tilde{q}^{*}=q\odot q^{*}\boxplus\epsilon(q\odot q'^{*}+q'\odot q^{*})=q_{I}\boxplus\epsilon\vec{0}
\]
This implies $||q||=1$ and $q\odot q'^{*}+q'\odot q^{*}=\vec{0}$,
which are the two unity constraints for dual quaternions. The inverse
of a unit dual quaternion is mathematically defined as 
\begin{equation}
	\tilde{q}^{-1}=(q\boxplus\epsilon q')^{-1}=q^{*}\boxminus\epsilon(q^{*}\odot q'\odot q^{*})\label{DQ_inv}
\end{equation}
similar to quaternions, the inverse and conjugate coincide for unit
dual quaternions. A unit dual quaternion can be used to represent
a rigid body transformation given by 
\[
T=\begin{bmatrix}R & t\\
	0_{1\times3} & 1
\end{bmatrix}\in\mathbb{SE}(3)
\]
This transformation encodes a rotation $R\in\mathbb{SO}(3)$ followed
by a translation $t\in\mathbb{R}^{3}$. Let $\tilde{q}_{R}=q\boxplus\epsilon\vec{0}$
be a dual quaternion for a pure rotation and $\tilde{q}_{t}=q_{I}\boxplus\epsilon\frac{1}{2}\vec{t}$
be a unit dual quaternion for the translation, where $\vec{t}=[0,t^{\top}]^{\top}\in\mathbb{R}^{4}$.
Then, the dual quaternion representation for the transformation $T$
has the following form 
\begin{equation}
	\tilde{q}=\tilde{q}_{t}\otimes\tilde{q}_{R}=q\boxplus\epsilon(\frac{\vec{t}\odot q}{2})\label{DQ_RBT}
\end{equation}
From \eqref{DQ_RBT} $q'=\frac{\vec{t}\odot q}{2}$ which can be used
to recover the translation vector using the relation 
\begin{equation}
	\vec{t}=2q'\odot q^{*}\label{recov_transl}
\end{equation}

\subsection{Twistors}

\textcolor{black}{To represent rigid-body motion in a minimal form,
	a six-dimensional twistor parameterization is employed. While unit
	dual quaternions provide a globally valid representation of pose,
	they are constrained and therefore not directly suitable for uncertainty
	representation in filtering frameworks. A twistor is defined as 
	\begin{equation}
		\tilde{\tau}=\frac{\tilde{q}-\tilde{q}_{I}}{\tilde{q}+\tilde{q}_{I}}=(\tilde{q}-\tilde{q}_{I})\otimes(\tilde{q}+\tilde{q}_{I})^{-1}\label{dq_2_twist}
	\end{equation}
	The twistor can be expressed in dual form as 
	\begin{equation}
		\tilde{\tau}=[0\hspace{2mm}\mu]\boxplus\epsilon[0\hspace{2mm}\rho]\label{dual_twistor}
	\end{equation}
	where $\mu$ is the MRP corresponding to the rotation quaternion $q$
	defined as 
	\[
	\mu=\frac{q_{v}}{1+q_{0}}\in\mathbb{R}^{3}
	\]
	and $\rho$ is the translational component of the twistor defined
	by 
	\[
	\rho=\Psi t\in\mathbb{R}^{3}
	\]
	with $t\in\mathbb{R}^{3}$ being the translation vector associated
	with rigid-body transformation and the matrix $\Psi$ is given by
	\begin{equation}
		\Psi=\frac{1}{4}(1-\mu^{\top}\mu)I_{3}-\frac{1}{2}[\mu]_{\times}+\frac{1}{2}\mu\mu^{\top}\in\mathbb{R}^{3\times3}
	\end{equation}
	The inverse mapping from the twistor to the corresponding unit dual
	quaternion is obtained from \eqref{dq_2_twist} as 
	\begin{equation}
		\tilde{q}=\frac{\tilde{q}_{I}+\tilde{\tau}}{\tilde{q}_{I}-\tilde{\tau}}\in\mathbb{R}^{8}\label{twist_2_dq}
	\end{equation}
	from \eqref{dual_twistor}, the twistor contains six independent parameters
	and the normalization constraint of $\tilde{q}$ is satisfied for
	any value of $\tilde{\tau}$ 
	\begin{equation}
		\tilde{q}^{*}\tilde{q}=\left(\frac{\tilde{q}_{I}+\tilde{\tau}}{\tilde{q}_{I}-\tilde{\tau}}\right)^{*}\left(\frac{\tilde{q}_{I}+\tilde{\tau}}{\tilde{q}_{I}-\tilde{\tau}}\right)=\left(\frac{\tilde{q}_{I}-\tilde{\tau}}{\tilde{q}_{I}+\tilde{\tau}}\right)\left(\frac{\tilde{q}_{I}+\tilde{\tau}}{\tilde{q}_{I}-\tilde{\tau}}\right)=\tilde{q}_{I}
	\end{equation}
}

\section{Dual Quaternion-based UKF\label{sec:Dual-Quaternion-based-UKF}}

The UKF leverages the concept of the Unscented Transform (UT) to approximate
the mean and covariance of a random variable undergoing a nonlinear
transformation \cite{Julier1997}. Given a random variable $C$ with
a known distribution and applying a nonlinear transformation $f_{\text{nl}}(\cdot)$,
the resulting approximated random variable $D$ and its distribution
can be obtained 
\begin{equation}
	D=f_{\text{nl}}(C)\label{UT}
\end{equation}
The transform in \eqref{UT} works by selecting a set of sample points,
known as sigma points, from the initial known distribution. The sigma
points are chosen in a specific way to represent the spread (mean
and covariance) of the state's probability distribution. These sigma
points are then passed through the nonlinear system model. After this
transformation, the UKF calculates a new mean and covariance based
on the transformed sigma points. This process provides an approximation
of the priori distribution.

\subsection{Filter Initialization}

The proposed DQUKF framework is based on the conventional UKF \cite{Haykin2004}
with certain modifications in specific areas that enable efficient
dual quaternion handling. In the Bayesian filtering framework, the
UKF recursively estimates the posterior probability distribution given
as 
\begin{equation}
	p(x_{k}|z_{k})=\frac{p(z_{k}|x_{k})p(x_{k}|z_{k-1})}{p(z_{k}|z_{k-1})}\label{bayesian}
\end{equation}
where the predicted \textit{a priori} distribution and marginal likelihood
in \eqref{bayesian} are defined as 
\begin{equation}
	\begin{cases}
		p(x_{k}|z_{k-1}) & =\int p(x_{k}|x_{k-1})p(x_{k-1}|z_{k-1})dx_{k-1}\\
		p(z_{k}|z_{k-1}) & =\int p(z_{k}|x_{k})p(x_{k}|z_{k-1})dx_{k}
	\end{cases}\label{integrals}
\end{equation}
calculating the integrals in \eqref{integrals} is not possible, they
can only be estimated. To follow the standard UKF notations, let the
estimated posterior and priori distribution $p(x_{k}|z_{k})$ and
$p(x_{k}|z_{k-1})$ be $\hat{x}_{k|k}$ and $\hat{x}_{k|k-1}$, respectively.
Let the estimated marginal likelihood $p(z_{k}|z_{k-1})$ be $\hat{z}_{k|k-1}$.
Before starting the filter, the state vector or parameters being estimated
should be clearly known. For navigation purposes the state vector
can be defined as 
\begin{equation}
	x=\begin{bmatrix}\tilde{q}^{\top} & \text{v}^{\top} & b_{\omega}^{\top} & b_{\alpha}^{\top}\end{bmatrix}^{\top}\in\mathbb{R}^{17}\label{state}
\end{equation}
where $\tilde{q}\in\mathbb{R}^{8}$ is the nominal dual quaternion
pose encoding both orientation and position, $\text{v}\in\mathbb{R}^{3}$
is the linear velocity, $b_{\omega}\in\mathbb{R}^{3}$ and $b_{\alpha}\in\mathbb{R}^{3}$
are the gyroscope and accelerometer biases, respectively. Although
the nominal pose is represented by the unit dual quaternion, the UKF
is not applied directly to the 8-dimensional dual quaternion coordinates.
Instead, the filter is formulated in an error-state manner. The nominal
state is propagated in dual quaternion form, whereas the associated
uncertainty is represented in a 6-dimensional local twistor coordinate.
Consequently, sigma point generation, covariance propagation, and
measurement correction are all performed in the local error space,
while the nominal pose remains on the unit dual quaternion manifold.
To start the filter, the state and covariance initial estimates need
to be established. To initialize the filter, \textcolor{black}{let
	the initial state estimate be defined as 
	\begin{equation}
		\hat{x}_{0|0}=\begin{bmatrix}\hat{\tilde{q}}_{0}^{\top} & \hat{\text{v}}_{0}^{\top} & \hat{b}_{\omega_{0}}^{\top} & \hat{b}_{\alpha_{0}}^{\top}\end{bmatrix}^{\top}\in\mathbb{R}^{17}\label{xhat0}
	\end{equation}
	The initial pose and velocity are obtained by perturbing the ground
	truth with deliberately large errors, including a significant attitude
	misalignment, a position error of $3.46~\mathrm{m}$, and a velocity
	error of $0.37~\mathrm{m/s}$ in order to demonstrate the convergence
	capability of the proposed filter under significant initialization
	uncertainty. Such initialization conditions are consistent with large
	misalignment scenarios commonly encountered in inertial navigation
	systems, where attitude errors exceeding several degrees invalidate
	small-angle assumptions and challenge linearized estimation methods
	\cite{Zeyuan2026_large_misalignment}.}The initial nominal pose dual
quaternion is constructed from the initial attitude and position estimates
using the rigid-body transformation relation in \eqref{DQ_RBT}, rather
than by assigning the dual part directly, which establishes the coupling
between the rotation and translation components. The state covariance
represents the current uncertainty in the state estimate. The large
values indicate that the filter is uncertain of the estimate. The
covariance is generally computed based on the deviation of the nominal
state from the mean value. This deviation in dual quaternions is defined
by the relative change in pose. Defining the side dependent subtraction
operator $\ominus$ and utilizing the relations in \eqref{Dq_multiply},\eqref{DQ_inv}
the pose error is given by 
\begin{equation}
	\delta\tilde{q}=\tilde{q}_{1}\ominus\tilde{q}_{2}=\tilde{q}_{1}\otimes\tilde{q}_{2}^{-1}\in\mathbb{R}^{8}\label{dq-dq}
\end{equation}
Equation \eqref{dq-dq} represents the relative error between two
dual quaternions $\tilde{q}_{1}$ and $\tilde{q}_{2}$. In this error
dual quaternion the real part is attitude error and the dual part
is the translational error. Moreover, dual quaternions are represented
by eight components, but due to both orthogonality and unit-norm constraints,
it effectively encodes only six independent degrees of freedom. The
relative pose between two nominal dual quaternions is first computed
as a unit dual quaternion error. However, this error is not used directly
in the covariance recursion. Instead, it is mapped to the corresponding
6-dimensional twistor coordinates, which provide the local Euclidean
coordinates used for covariance representation and sigma point generation.
The goal of the filter is to estimate the state vector in \eqref{state}
while handling dual quaternions properly within the recursive framework.
\textcolor{black}{The initial covariance associated with the state
	vector \eqref{xhat0} and takes into account the reduced dimensionality
	is given by 
	\begin{equation}
		P_{0|0}=\begin{bmatrix}\eta_{1}I_{6} & 0_{6\times3} & 0_{6\times6}\\
			0_{3\times6} & \eta_{2}I_{3} & 0_{3\times6}\\
			0_{6\times6} & 0_{6\times3} & \eta_{3}I_{6}
		\end{bmatrix}\in\mathbb{R}^{15\times15}\label{P0}
	\end{equation}
	Where $\eta_{1},\eta_{2},\eta_{3}$ are empirically tuned constants
	for the initial state covariance. There are also two additional covariance
	matrices, namely, the process noise and measurement noise covariance
	matrices, which models the uncertainty in the system dynamics and
	measurements, respectively. 
	\begin{equation}
		R=\mu*I_{3n_{x}}\in\mathbb{R}^{3n_{x}\times3n_{x}}\label{R_matrix}
	\end{equation}
	\begin{equation}
		Q=\begin{bmatrix}0_{9\times9} & 0_{9\times3} & 0_{9\times3}\\
			0_{3\times9} & Q_{\omega} & 0_{3\times3}\\
			0_{3\times9} & 0_{3\times3} & Q_{\alpha}
		\end{bmatrix}\in\mathbb{R}^{15\times15}\label{Q_matrix}
	\end{equation}
	where 
	\[
	Q_{\omega}=\sigma_{b_{\omega}^{2}}I_{3}\hspace{3mm}\text{and}\hspace{3mm}Q_{\alpha}=\sigma_{b_{\alpha}^{2}}I_{3}
	\]
	where $\mu,\sigma_{b_{\omega}},\sigma_{b_{\alpha}}$ are tuning parameters
	for the measurement and process noise covariances. The parameters
	for the process noise covariance are obtained via hardware specifications
	such as spectral densities of the sensors provided in the dataset.
	$Q_{\omega}$ and $Q_{\alpha}$ represent the uncertainty in the bias
	evolution for the angular velocity and acceleration respectively.
	These covariance matrices are essential for the calculation of the
	state estimate and the covariance associated with the predicted measurements.
	The measurement prediction is performed within the VIO framework,
	hence the associated covariance relies on the initialized measurement
	noise covariance matrix in \eqref{R_matrix}. The dimension of the
	measurement covariance assumes that there are $n_{x}$ landmarks observed
	in every frame and each landmark has a $x$, $y$ and $z$ coordinate.
	From \eqref{Q_matrix}, it is also clear that the uncertainty in evolution
	of the pose and velocity is zero. The dimension of the process noise
	covariance matrix reflects the reduced dimensionality formulated via
	twistors. There is also an additional process covariance to account
	for sensor noise. 
	\begin{equation}
		Q_{n}=\begin{bmatrix}\text{diag}(\bar{b}_{\omega}) & 0_{3\times3}\\
			0_{3\times3} & \text{diag}(\bar{b}_{\alpha})
		\end{bmatrix}\in\mathbb{R}^{6\times6}\label{Q_n}
	\end{equation}
	where 
	\[
	\bar{b}_{\omega}=\sigma_{g}I_{3}\hspace{3mm}\text{and}\hspace{3mm}\bar{b}_{\alpha}=\sigma_{a}I_{3}
	\]
	$\sigma_{g}$ and $\sigma_{a}$ are scaling parameters for the mean
	gyroscope and accelerometer noise, respectively. These parameters
	are obtained from the IMU hardware specification provided for the
	dataset.} Following the initialization of the state in \eqref{xhat0}
and the corresponding covariance in \eqref{P0}, the sigma points
are generated. The UKF heavily relies on the augmented state and covariance
matrix for sigma point generation. \textcolor{black}{Sigma points are
	generated in the twistor parameterization of the local error state,
	as it provides a minimal, unconstrained representation that avoids
	the normalization constraints associated with dual quaternions. This
	enables the unscented transform to be applied in a Euclidean space.
	The resulting sigma points are then mapped back to the nominal state
	using the relation in \eqref{twist_2_dq} to obtain the corresponding
	sigma points in the state space. The local twistor error state can
	be defined as 
	\begin{equation}
		\delta x_{k}=\begin{bmatrix}\delta\tau_{k}^{\top} & \delta\text{v}_{k}^{\top} & \delta b_{\omega,k}^{\top} & \delta b_{\alpha,k}^{\top}\end{bmatrix}\in\mathbb{R}^{15}\label{err_state}
	\end{equation}
} the local error state is initialized as $\delta x_{0|0}=0_{1\times15}$.
IMU readings are noisy and this noise should be modeled as a part
of the state vector alongside the IMU biases. Let's define the noise
vector as 
\begin{equation}
	x_{n}=\begin{bmatrix}n_{\omega,k}^{\top} & n_{\alpha,k}^{\top}\end{bmatrix}^{\top}\in\mathbb{R}^{6}\label{noise}
\end{equation}
The augmented state vector is formed by combining the local error-state
vector in \eqref{err_state}, rather than the nominal state, with
the expected value of the noise vector in \eqref{noise}, which is
zero under the assumption of white Gaussian noise 
\begin{equation}
	\delta\hat{x}_{k-1|k-1}^{a}=\left[\delta x_{k-1|k-1}^{\top},\hspace{2mm}0_{6\times1}^{\top}\right]^{\top}\in\mathbb{R}^{21}\label{x_aug}
\end{equation}
In developing the filter, the initial gyroscope and accelerometer
bias estimates are initialized from sample averages, after which both
biases are treated as filter states and recursively estimated. For
the augmented state covariance, the matrices in \eqref{P0} and \eqref{Q_n}
are diagonalized as follows 
\begin{equation}
	P_{k-1|k-1}^{a}=\begin{bmatrix}P_{k-1|k-1} & 0_{15\times6}\\
		0_{6\times15} & Q_{n}
	\end{bmatrix}\in\mathbb{R}^{21\times21}\label{p_aug}
\end{equation}
Both the augmented state and covariance are expressed in the same
minimal representation, ensuring that uncertainty is consistently
defined and propagated within the local error-state space.

\subsection{Sigma Point Generation}

In the standard UKF, the number of sigma points depends on the dimension
of the augmented state vector. The total number of sigma points is
given by the following relation 
\begin{equation}
	n=2L+1\label{no.sigma}
\end{equation}
where $L$ is the dimension of the augmented state vector. The total
number of sigma points becomes $n=43$. There are also additional
parameters that serve as arbitrary constants to determine suitable
characteristics for the generated sigma points. These parameters can
impact the filter performance and must be chosen wisely. These parameters
are defined as follows: $\alpha$ is the spread of sigma points, $\kappa$
is the secondary scaling parameter and $\beta$ incorporates prior
knowledge. In the designed filter, the exact values of these parameters
are 
\begin{equation}
	\alpha=10^{-4},\hspace{5mm}\kappa=0\hspace{5mm}\text{and}\hspace{5mm}\beta=2\label{params}
\end{equation}
The parameters in \eqref{params} are used in calculating important
quantities needed for weight calculation and sigma point generation.
\begin{equation}
	\lambda=3-L\hspace{5mm}\text{and}\hspace{5mm}\gamma=\sqrt{L+\lambda}\label{lambda-gamma}
\end{equation}
The weights control how much each sigma point contributes to the predicted
mean and covariance. There are two weighting constants for every sigma
point. The first is the mean weight $(W_{i}^{m})^{\top}\in\mathbb{R}^{n}$
and the second is the covariance weight $(W_{i}^{c})^{\top}\in\mathbb{R}^{n}$.
These weights are defined in terms of the parameters in \eqref{lambda-gamma}
\begin{equation}
	W_{0}^{m}=\frac{\lambda}{L+\lambda}\hspace{5mm}\text{and}\hspace{5mm}W_{i}^{m}=\frac{1}{2(L+\lambda)}\label{Wm}
\end{equation}
where $i=1,2,3\ldots n-1$. The covariance weights are defined in
terms of parameters in \eqref{params} and \eqref{lambda-gamma} as
\begin{equation}
	W_{0}^{c}=\frac{\lambda}{L+\lambda}+(1-\alpha^{2}+\beta)\hspace{5mm}\text{and}\hspace{5mm}W_{i}^{c}=W_{i}^{m}\label{Wc}
\end{equation}
The sigma points are generated around the mean using the following
relation 
\begin{equation}
	\mathcal{X}_{k-1|k-1}^{a}=\{\bar{x},\hspace{2mm}\bar{x}+\gamma\sqrt{P^{a}},\hspace{2mm}\bar{x}-\gamma\sqrt{P^{a}}\}
\end{equation}
Substituting the value of $\gamma$ from \eqref{lambda-gamma} and
the mean in \eqref{x_aug}, the expression for the sigma points can
be rewritten as 
\begin{equation}
	\begin{cases}
		\mathcal{X}_{k-1|k-1}^{a,(0)} & =\delta\hat{x}_{k-1|k-1}^{a}\\
		\mathcal{X}_{k-1|k-1}^{a,(j)} & =\delta\hat{x}_{k-1|k-1}^{a}+\sqrt{(L+\lambda)P_{k-1|k-1}^{a}}\\
		\mathcal{X}_{k-1|k-1}^{a,(j+L)} & =\delta\hat{x}_{k-1|k-1}^{a}-\sqrt{(L+\lambda)P_{k-1|k-1}^{a}}
	\end{cases}\in\mathbb{R}^{43\times21}\label{sigmapts}
\end{equation}
where $j=1,\ldots,L$. To find the square root quantity in \eqref{sigmapts}
two common approaches are used. The first approach is using the Cholesky
decomposition to obtain a lower triangular matrix that satisfies the
following 
\begin{equation}
	(L+\lambda)P_{k-1|k-1}^{a}=BB^{\top}\label{Cholesky}
\end{equation}
Equation \eqref{Cholesky} is only valid if the scaled augmented covariance
matrix under the square root is always symmetric positive definite.
The second approach is using Singular Value Decomposition (SVD) to
obtain the square root of a matrix. SVD decomposes the matrix as follows.
\begin{equation}
	(L+\lambda)P_{k-1|k-1}^{a}=USV^{\top}\label{svd}
\end{equation}
where $U$, $S$ and $V$ are matrices of the left singular vectors,
singular values and right singular vectors, respectively. The SVD
method only works if the matrix being decomposed is symmetric positive
semi-definite. This means that the matrices consisting of right and
left singular vectors are identical. Then \eqref{svd} can be written
as 
\[
(L+\lambda)P_{k-1|k-1}^{a}=USU^{\top}
\]
where $UU^{\top}=I$, hence 
\begin{equation}
	\sqrt{(L+\lambda)P^{a}}=U\sqrt{S}\hspace{3mm}\in\mathbb{R}^{21\times21}\label{svd_decomp}
\end{equation}
In this paper, \eqref{svd_decomp} is used for finding the square
root as it does not strictly require the scaled augmented covariance
to be positive definite and provides better numerical stability. \textcolor{black}{Each
	sigma point is a 21-dimensional perturbation vector expressed entirely
	in Euclidean coordinates. Therefore, the addition and subtraction
	operations used in sigma point generation are standard vector operations
	and do not involve dual quaternion algebra especially since the perturbations
	produced by the square root of the covariance in \eqref{svd_decomp}
	are already expressed in the same Euclidean space. 
	\begin{equation}
		\begin{aligned}\mathcal{X}_{k-1|k-1}^{a,(i)}=\Big[ & \delta\tau_{k-1|k-1}^{(i)\top},\;\delta\mathbf{v}_{k-1|k-1}^{(i)\top},\;\delta b_{\omega,k-1|k-1}^{(i)\top},\\
			& \delta b_{\alpha,k-1|k-1}^{(i)\top},\;n_{\omega,k-1|k-1}^{(i)},\;n_{\alpha,k-1|k-1}^{(i)}\Big]^{\top}\in\mathbb{R}^{21}
		\end{aligned}
	\end{equation}
	where $i=0,\ldots,2L$. The resulting sigma points perturbations are
	subsequently mapped to the nominal state through dual quaternion composition.
	\begin{equation}
		\delta\tilde{q}_{k-1|k-1}^{(i)}=\mathcal{T}^{-1}(\delta\tau_{k-1|k-1}^{(i)})\label{dq_pert}
	\end{equation}
	where $\mathcal{T}^{-1}(\cdot)$ denotes the inverse twistor to unit
	dual quaternion map given in \eqref{twist_2_dq}. The nominal pose
	is obtained by using \eqref{dq-dq} and \eqref{dq_pert} as 
	\begin{equation}
		\tilde{q}_{k-1|k-1}^{(i)}=\hat{\tilde{q}}_{k-1|k-1}\otimes\delta\tilde{q}_{k-1|k-1}^{(i)}\label{pose_nom}
	\end{equation}
	where $\hat{\tilde{q}}_{k-1|k-1}$ is obtained from initial state
	estimate in \eqref{xhat0} for the first iteration. The resulting
	nominal state sigma points can be separated into state and measurement
	noise components as follows. 
	\begin{equation}
		\mathcal{X}_{k-1|k-1}^{a}=\begin{bmatrix}\mathcal{X}_{k-1|k-1}^{x} & \mathcal{X}_{k-1|k-1}^{\omega} & \mathcal{X}_{k-1|k-1}^{\alpha}\end{bmatrix}^{\top}\in\mathbb{R}^{43\times23}
	\end{equation}
} where $\mathcal{X}_{k-1|k-1}^{x}=[\mathcal{X}_{k-1|k-1}^{\tilde{q}},\mathcal{X}_{k-1|k-1}^{\text{v}},\mathcal{X}_{k-1|k-1}^{b_{\omega}},\mathcal{X}_{k-1|k-1}^{b_{\alpha}}]\in\mathbb{R}^{43\times17}$
are the sigma points corresponding to the state $\mathcal{X}_{k-1|k-1}^{\omega}\in\mathbb{R}^{43\times3}$
are the sigma points for the gyroscope noise and $\mathcal{X}_{k-1|k-1}^{\alpha}\in\mathbb{R}^{43\times3}$
are those for accelerometer noise. This sectioning is useful when
propagating the sigma points.

\subsection{Dual Quaternion Kinematics}

For each sigma point, only the nominal state components are propagated
through the nonlinear process model. The sigma-point components associated
with IMU noise are not propagated as independent states. Instead,
they are used within the same time step to form corrected gyroscope
and accelerometer inputs for that sigma point. Only those associated
with the state vector $\mathcal{X}_{k-1|k-1}^{x}$ are evolved according
to the system dynamics. The dynamics are driven by IMU measurements
that have been corrected for sensor bias and noise, achieved by subtracting
the estimated bias and noise from both gyroscope and accelerometer
readings. 
\begin{equation}
	\begin{cases}
		\omega_{k}^{(i)} & =\omega_{m,k}-b_{\omega,k}^{(i)}-n_{\omega,k}^{(i)}\\
		a_{k}^{(i)} & =a_{m,k}-b_{\alpha,k}^{(i)}-n_{\alpha,k}^{(i)}
	\end{cases}\label{corr_meas}
\end{equation}
\textcolor{black}{The accuracy of the propagation step is dependent
	on the quality of the inertial measurements and the fidelity of the
	bias and noise modeling. Since the corrected inputs are obtained by
	subtracting estimated bias and noise terms from raw IMU measurements,
	any residual bias or high noise density directly affects the state
	evolution and leads to increased uncertainty during prediction. High-grade
	inertial sensors, such as those used in aerospace navigation systems,
	exhibit significantly lower noise densities and bias instability,
	resulting in more accurate state propagation and reduced drift \cite{Zeyuan2021}.
	In contrast, low-cost IMUs introduce larger stochastic errors, which
	accumulate over time and degrade estimation accuracy unless sufficiently
	corrected by measurement updates.} The biases and noise terms in \eqref{corr_meas}
are obtained from the generated sigma instead of the random walk model.
\begin{equation}
	\begin{split}b_{\omega}=\mathcal{X}_{k-1|k-1}^{x,b_{\omega}},\hspace{3mm}n_{\omega}=\mathcal{X}_{k-1|k-1}^{\omega}\\
		b_{\alpha}=\mathcal{X}_{k-1|k-1}^{x,b_{\alpha}},\hspace{3mm}n_{\alpha}=\mathcal{X}_{k-1|k-1}^{\alpha}
	\end{split}
\end{equation}
Let $\mathcal{X}_{k-1|k-1}^{x,\tilde{q}}=[\mathcal{X}_{k-1|k-1}^{x,q},\mathcal{X}_{k-1|k-1}^{x,q'}]\in\mathbb{R}^{43\times8}$.
For this filter, the continuous-time system dynamics are defined as
follows \cite{Sveier2021}. 
\begin{equation}
	\begin{cases}
		\tilde{\dot{q}}_{\mathcal{B}/\mathcal{W}} & =\frac{1}{2}\tilde{q}_{\mathcal{B}/\mathcal{W}}\otimes\tilde{\omega}_{\mathcal{B}/\mathcal{W}}^{\mathcal{B}}\\
		\dot{\text{v}} & =g+\mathcal{R}_{q}(\mathcal{X}_{k-1|k-1}^{x,q_{k}^{(i)}})a_{k}^{(i)}\in\mathbb{R}^{3}
	\end{cases}\label{dynamics}
\end{equation}
where $\tilde{q}_{\mathcal{B}/\mathcal{W}}$ is dual quaternion representation
for the motion in the body frame, $\tilde{\omega}_{\mathcal{B}/\mathcal{W}}^{\mathcal{B}}=\omega_{\mathcal{B}/\mathcal{W}}^{\mathcal{B}}\boxplus\epsilon\text{v}_{\mathcal{B}/\mathcal{W}}^{\mathcal{B}}$
is the dual vector for the twist in the body frame and $\mathcal{R}_{q}$
is the rotation matrix corresponding to the real part of the dual
quaternion. The angular velocity is obtained from the IMU measurements
in \eqref{corr_meas} and the linear velocity comes from the state
sigma points in $\mathcal{X}_{k-1|k-1}^{x,\text{v}}$. Since the filter
operates in discrete time, it is essential to discretize the continuous
dynamics. For a displacement given by a constant twist $\tilde{\omega}$
within a time interval $\Delta t$, the screw motion $\tilde{\theta}\tilde{k}=\Delta t\tilde{\omega}$
of this displacement is given by the unit dual quaternion 
\[
\tilde{q}=\text{exp}(\frac{\Delta t\tilde{\omega}_{k}}{2})
\]
Thus, for the given system, the discrete time dynamics can written
as 
\begin{equation}
	\begin{cases}
		\tilde{q}_{k+1}^{(i)} & =\exp(\frac{\Delta t\tilde{\omega}_{k}}{2})\tilde{q}_{k}^{(i)}\\
		\text{v}_{k+1}^{(i)} & =\text{v}_{k}^{(i)}+\Delta t(g+\mathcal{R}_{q}(\mathcal{X}_{k-1|k-1}^{x,q_{k}^{(i)}})a_{k}^{(i)})
	\end{cases}\label{eq:discrete_dynamics}
\end{equation}
To solve the discrete dual quaternion dynamics neatly in matrix form
without dealing with the exponential of the dual vector $\tilde{\omega}$,
consider the following 
\[
\Gamma(\omega)=\begin{bmatrix}0 & -\omega_{k}^{(i)\top}\\
	\omega_{k}^{(i)} & -[\omega_{k}^{(i)}]_{\times}
\end{bmatrix},\hspace{3mm}\Gamma(\text{v})=\begin{bmatrix}0 & -(\text{v}_{k}^{\mathcal{B}})^{\top}\\
	\text{v}_{k}^{\mathcal{B}} & -[\text{v}_{k}^{\mathcal{B}}]_{\times}
\end{bmatrix}\in\mathbb{R}^{4\times4}
\]
where $\text{v}_{k}^{\mathcal{B}}=\mathcal{R}_{q}(\mathcal{X}^{q_{k}^{(i)}})^{\top}\text{v}_{k}^{(i)}$.
Using the previous matrices, the combined matrix is given by 
\begin{equation}
	M_{k}^{(i)}=\begin{bmatrix}\Gamma(\omega_{k}^{(i)}) & 0_{4\times4}\\
		\Gamma(\text{v}_{k}^{\mathcal{B}}) & \Gamma(\omega_{k}^{(i)})
	\end{bmatrix}\in\mathbb{R}^{8\times8}\label{exp_matrix}
\end{equation}
The matrix in \eqref{exp_matrix} can be used instead of the dual
vector, resulting in the following simplified expression 
\begin{equation}
	\tilde{q}_{k+1}^{(i)}=\exp(\frac{\Delta t}{2}M_{k}^{(i)})\tilde{q}_{k}^{(i)}\label{mat_exp}
\end{equation}
In \eqref{eq:discrete_dynamics}, the gravity vector is given by $g=[0,0,-9.81]^{\top}$
and the time step used for propagating the sigma points through the
system dynamics is $\Delta t=0.05$ s. The output of propagating each
sigma point through the process model is a new state vector 
\begin{equation}
	\hat{x}_{k|k-1}=\begin{bmatrix}\tilde{q}_{k|k-1}^{\top} & \text{v}_{k|k-1}^{\top} & b_{\omega}^{\top} & b_{\alpha}^{\top}\end{bmatrix}\in\mathbb{R}^{17}
\end{equation}
repeating the same process for all 43 sigma points produces the propagated
sigma point matrix for the current iteration. 
\begin{equation}
	\mathcal{X}_{k|k-1}^{a}=f(\mathcal{X}_{k-1|k-1}^{a},u_{k-1})\in\mathbb{R}^{43\times17}\label{propagated_sigma_pts}
\end{equation}
The next step is to use these propagated sigma points to find the
predicted state and covariance matrices. The \textit{a priori} state
estimate is obtained by multiplying the sigma points with the corresponding
weights from \eqref{Wm}. This is done as follows: 
\begin{equation}
	\begin{split}\hat{x}_{k|k-1} & =\sum_{i=0}^{n-1}W_{i}^{m}*f(\mathcal{X}_{k-1|k-1}^{a},u_{k-1})\\
		& =\sum_{i=0}^{n-1}W_{i}^{m}*\mathcal{X}_{k|k-1}^{a}\in\mathbb{R}^{17}
	\end{split}
	\label{weighted_avg}
\end{equation}
\textcolor{black}{Since the propagated sigma points contain dual quaternion
	elements. The weighted average in \eqref{weighted_avg} can be done
	for dual quaternions by setting the first propagated sigma points
	as the reference and computing the relative pose for the remaining
	sigma point with respect to this reference. 
	\begin{equation}
		\tilde{q}_{k|k-1}^{\text{ref}}=\tilde{q}_{k|k-1}^{(0)}\label{dq_ref}
	\end{equation}
	using \eqref{dq_ref} the relative pose can be calculated as follows
	\begin{equation}
		\delta\tilde{q}_{k|k-1}^{(i)}=\tilde{q}_{k|k-1}^{\text{ref}^{-1}}\otimes\tilde{q}_{k|k-1}^{(i)}
	\end{equation}
	After obtaining the relative dual quaternions, they are mapped to
	twistor representations. 
	\begin{equation}
		\delta\tilde{\tau}_{k|k-1}^{(i)}=\mathcal{T}(\delta\tilde{q}_{k|k-1}^{(i)})
	\end{equation}
	The averaging for the twistor can then be done in the Euclidean space
	as follows 
	\begin{equation}
		\delta\tau_{k|k-1}^{\text{avg}}=\sum_{i=0}^{n-1}W_{i}^{m}\delta\tilde{\tau}_{k|k-1}^{(i)}
	\end{equation}
	The average twistor then needs to be mapped back to global pose represented
	via dual quaternions. 
	\[
	\delta q_{k|k-1}^{\text{avg}}=\mathcal{T}^{-1}(\delta\tau_{k|k-1}^{\text{avg}})
	\]
	then the predicted pose is given by 
	\begin{equation}
		\hat{\tilde{q}}_{k|k-1}=\tilde{q}_{k|k-1}^{\text{ref}}\otimes\delta q_{k|k-1}^{\text{avg}}
	\end{equation}
} The remaining states do not require this special handling and can
be averaged algebraically 
\begin{equation}
	\begin{cases}
		\hat{\text{v}}_{k|k-1} & =\sum_{i=0}^{n-1}W_{i}^{m}\text{v}_{k|k-1}^{(i)}\\
		\hat{b}_{\omega,k|k-1} & =\sum_{i=0}^{n-1}W_{i}^{m}b_{\omega}^{(i)}\\
		\hat{b}_{\alpha,k|k-1} & =\sum_{i=0}^{n-1}W_{i}^{m}b_{\alpha}^{(i)}
	\end{cases}\label{non_dual_avg}
\end{equation}
The \textit{a priori} state estimate is then defined as 
\begin{equation}
	\hat{x}_{k|k-1}=\begin{bmatrix}\hat{\tilde{q}}_{k|k-1}^{\top} & \hat{\text{v}}_{k|k-1}^{\top} & \hat{b}_{\omega,k|k-1}^{\top} & \hat{b}_{\alpha,k|k-1}^{\top}\end{bmatrix}^{\top}\in\mathbb{R}^{17}\label{pred_state}
\end{equation}
The \textit{a priori} covariance estimate can also be computed using
the weights in \eqref{Wc}, the propagated sigma point in \eqref{propagated_sigma_pts}
and the \textit{a priori} state estimate in \eqref{pred_state}. \textcolor{black}{The
	process covariance matrix in \eqref{Q_matrix} is also added to the
	predicted covariance to account for the uncertainty due to unmodeled
	process dynamics. 
	\begin{equation}
		P_{k|k-1}=\sum_{i=0}^{n-1}W_{i}^{c}*(\mathcal{X}_{k|k-1}^{a}-\hat{x}_{k|k-1})(\mathcal{X}_{k|k-1}^{a}-\hat{x}_{k|k-1})^{\top}+Q\label{P_pred}
	\end{equation}
	where $P_{k|k-1}\in\mathbb{R}^{15\times15}$ and $W_{i}^{c}$ is being
	applied element wise to each entry of matrix. The deviation between
	the propagated sigma points and the $\textit{a priori}$ state estimate
	in \eqref{P_pred} is performed in the local error space. The pose
	uncertainty is obtained by computing the relative transformation between
	the propagated sigma point and the predicted state, and mapping this
	relative dual quaternion to its corresponding twistor representation.
	The deviation for non-pose elements are computed using standard algebraic
	differences. The subtraction in \eqref{P_pred} can then be expressed
	as 
	\begin{equation}
		\mathcal{X}_{k|k-1}^{a}-\hat{x}_{k|k-1}=\begin{bmatrix}\delta\tilde{\tau}_{k|k-1}^{(i)}-\delta\tau_{k|k-1}^{\text{avg}}\\
			\text{v}_{k|k-1}^{(i)}-\hat{\text{v}}_{k|k-1}\\
			b_{\omega}^{(i)}-\hat{b}_{\omega,k|k-1}\\
			b_{\alpha}^{(i)}-\hat{b}_{\alpha,k|k-1}
		\end{bmatrix}\in\mathbb{R}^{15}\label{sigma-pred}
	\end{equation}
} The developed filter uses fixed landmarks as global reference points
to obtain measurements. This is an intuitive and simple approach used
in the standard UKF. These landmarks provide correction against sensor
drift, thereby improving the accuracy and consistency of the state
estimates. At the beginning of each run, $n_{x}$ landmarks are detected
with an upper bound of 60 landmarks. These landmarks are global anchors
and have $x,y$ and $z$ coordinates that can be structured in a matrix
form as 
\begin{equation}
	l_{k}=\begin{bmatrix}l_{1},l_{2},\ldots,l_{n_{x}}\end{bmatrix}=\begin{bmatrix}l_{1,x} & l_{2,x} & \ldots & l_{n_{x},x}\\
		l_{1,y} & l_{2,y} & \ldots & l_{n_{x},y}\\
		l_{1,z} & l_{2,z} & \ldots & l_{n_{x},z}
	\end{bmatrix}\in\mathbb{R}^{3\times n_{x}}\label{landmarks}
\end{equation}
The state vector contains the pose and velocity which are not directly
observable. These quantities are deduced from sensor measurements
like angular velocity (gyroscope) and acceleration (accelerometer).
These sensor measurements are indirect, noisy observations of these
states. The measurement model is used to relate the sensor measurements
to the state and is generally defined as 
\begin{equation}
	z_{k}=h(x_{k})+v_{k}
\end{equation}
where $h(x_{k})$ represent the nonlinear function that relates the
measurements to the states and $v_{k}$ is the sensor or measurement
noise. The fixed landmarks in \eqref{landmarks} are observed through
a camera, which provides a position-based observation, the nonlinear
function in the measurement model should be a function of orientation
and position since the camera measures the location of the landmark
relative to itself. This requires transforming the orientation from
the world frame to the camera frame and subtracting the camera position
from the landmark position. The measurement model representing this
transformation is given by 
\begin{equation}
	z_{k}=\text{vec}\left(\mathcal{R}_{q}(q_{k})^{\top}(l_{k}-p_{k})\right)\in\mathbb{R}^{3n_{x}\times1}\label{meas_model}
\end{equation}
where $q_{k}\in\mathbb{S}^{3}$ and $p_{k}\in\mathbb{R}^{3}$ denote
the ground truth attitude quaternion and position, respectively. Using
the landmarks defined in \eqref{landmarks}, the true measurement
vector can be obtained. In real world scenarios, the ground truth
is obtained via camera observation. For the predicted measurement
vector, the propagated sigma points are used instead of the ground
truth to acquire the measurement prediction. This is done by further
propagating the sigma points through the measurement model. 
\begin{equation}
	\mathcal{Z}_{k}=h(\mathcal{X}_{k|k-1}^{a},u_{k},\mathcal{X}_{k|k-1}^{a,v})\in\mathbb{R}^{43\times3n_{x}}
\end{equation}
it is worth noting that the propagated sigma points in $\mathcal{X}_{k|k-1}^{a}$
consist of dual quaternion, velocity and biases. It does not contain
any noise terms, which means that $\mathcal{X}_{k|k-1}^{a,v}=0$.
Then the measurement model is just a function of the sigma point corresponding
to the state, specifically $\mathcal{X}_{k|k-1}^{a,q}$ and $\mathcal{X}_{k|k-1}^{a,q'}$.
Let $\mathcal{Z}_{k}^{(i)}$ denote the $i$-th row of $\mathcal{Z}_{k}$,
the predicted measurement model can be written as 
\begin{equation}
	\mathcal{Z}_{k}^{(i)}=\text{vec}\left(\mathcal{R}_{q}(\mathcal{X}_{k|k-1}^{a,q^{(i)}})^{\top}(l_{k}-\mathcal{X}_{k|k-1}^{a,\text{t}^{(i)}})\right)\label{measurments}
\end{equation}
where $\mathcal{X}_{k|k-1}^{a,\text{t}^{(i)}}\in\mathbb{R}^{3}$ is
the position extracted from the dual part of the propagated sigma
points $\mathcal{X}_{k|k-1}^{a,q'^{(i)}}$. The predicted measurement
vector is obtained via weighted sum of the relation in \eqref{measurments}.
\begin{equation}
	\hat{z}_{k}=\sum_{i=0}^{n-1}W_{i}^{m}\mathcal{Z}_{k}^{(i)}\in\mathbb{R}^{(3*n_{x})\times1}\label{pred_measurement}
\end{equation}
Using the predicted measurement in \eqref{pred_measurement} it is
possible to compute the measurement covariance as follows 
\begin{equation}
	P_{k}^{z}=\sum_{i=0}^{n-1}W_{i}^{c}*(\mathcal{Z}_{k}^{(i)}-\hat{z}_{k})(\mathcal{Z}_{k}^{(i)}-\hat{z}_{k})^{\top}+R\in\mathbb{R}^{3n_{x}\times3n_{x}}\label{measurement_cov}
\end{equation}
where $R$ is the measurement noise covariance. It is also important
to establish the relation between the predicted states and predicted
measurements by computing the uncertainty state variables and camera
observations. This is done by finding the cross covariance matrix
defined as 
\begin{equation}
	P_{k|k-1}^{xz}=\sum_{i=0}^{n-1}W_{i}^{c}*(\mathcal{X}_{k|k-1}^{a}-\hat{x}_{k|k-1})(\mathcal{Z}_{k}^{(i)}-\hat{z}_{k})^{\top}\in\mathbb{R}^{15\times3n_{x}}\label{cross_cov}
\end{equation}

\begin{figure}[!ht]
	\centering{}\includegraphics[width=1\linewidth]{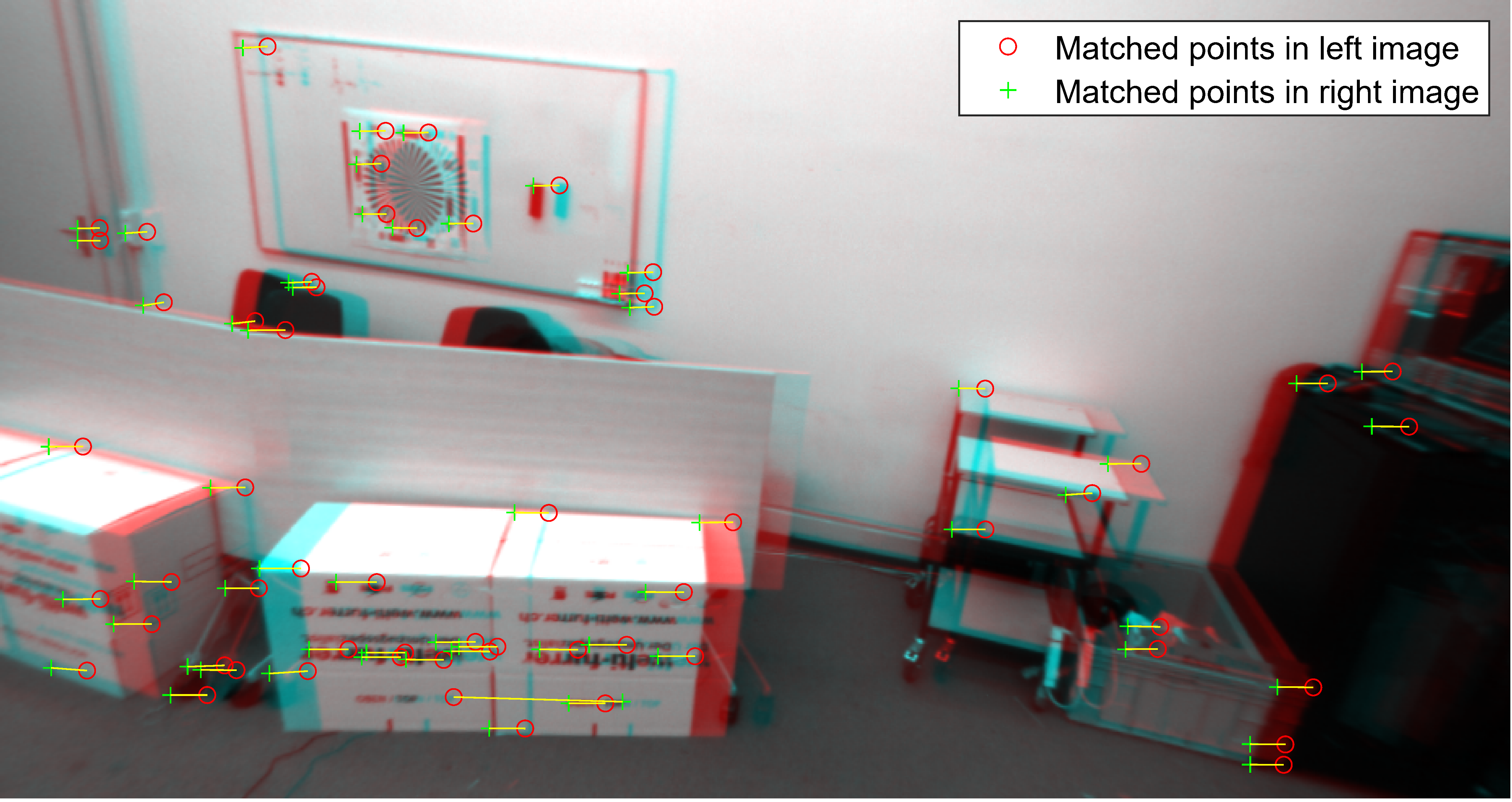} \caption{\label{feat_detect}Feature extraction and stereo matching using Shi-Thomasi
		on the EuRoC dataset.}
\end{figure}

Subtracting the \textit{a priori} state vector from the propagated
sigma points requires the same special handling as the one done in
\eqref{P_pred}. The measurement prediction $\hat{z}_{k}$ is subtracted
algebraically from the output of the measurement model in \eqref{measurments}.
In the update phase, the predicted state is corrected using the true
measurement obtained using the camera or ground truth in the case
of simulation. A useful quantity used in the update step is the innovation,
which represents the difference between the true and model predicted
measurements. Another quantity that is essential in all Kalman-type
filters is the Kalman gain. The Kalman gain controls the amount of
correction required based on the uncertainty present in the dynamical
and measurement models. If the uncertainty in the measurements is
high, the Kalman gain will be small, which means the filter trusts
the dynamical model more. The Kalman gain can be formally written
in terms of the measurement and cross covariances as follows 
\begin{equation}
	L_{k}=P_{k|k-1}^{xz}*(P_{k}^{z})^{-1}\in\mathbb{R}^{15\times3n_{x}}\label{kalman_gain}
\end{equation}
where $P_{k|k-1}^{xz}$ and $P_{k}^{z}$ are obtained from \eqref{cross_cov}
and \eqref{measurement_cov}, respectively. \textcolor{black}{Using
	the Kalman gain in \eqref{kalman_gain} and innovation, the state
	can be updated as follows 
	\begin{equation}
		\hat{x}_{k|k}=\hat{x}_{k|k-1}+L_{k}\underbrace{(z_{k}-\hat{z}_{k})}_{\text{Innovation}}\label{x_est}
	\end{equation}
	The addition is not straightforward since the innovation term is constructed
	using twistor representation while the }\textit{\textcolor{black}{a
		priori}}\textcolor{black}{{} state estimate consists of a dual quaternion.
	To resolve this issue the twistor is mapped to a dual quaternion perturbation
	using \eqref{twist_2_dq}. 
	\begin{equation}
		\psi_{k}=L_{k}(z_{k}-\hat{z}_{k})\in\mathbb{R}^{15}
	\end{equation}
	Let $\psi_{k}^{\tau}\in\mathbb{R}^{6}$ be the twistor for the pose
	innovation, then the addition of the dual quaternion elements in \eqref{x_est}
	can be done as follows 
	\[
	\delta\tilde{q}_{k}=\mathcal{T}^{-1}(\psi_{k}^{\tau})
	\]
	\begin{equation}
		\hat{\tilde{q}}_{k|k}=\hat{\tilde{q}}_{k|k-1}\otimes\delta\tilde{q}_{k}\label{upd_dq}
	\end{equation}
	The remaining non-dual quaternion elements are added algebraically.
	\begin{equation}
		\begin{cases}
			\hat{\text{v}}_{k|k} & =\hat{\text{v}}_{k|k-1}+\psi_{k}^{\text{v}}\\
			\hat{b}_{\omega,k|k} & =\hat{b}_{\omega,k|k}+\psi_{k}^{b_{\omega}}\\
			\hat{b}_{\alpha,k|k} & =\hat{b}_{\alpha,k|k}+\psi_{k}^{b_{\alpha}}
		\end{cases}\label{upd_non_dq}
	\end{equation}
} Combining the relation in \eqref{upd_dq} and \eqref{upd_non_dq},
the updated state is given by 
\[
\hat{x}_{k|k}=\begin{bmatrix}\hat{\tilde{q}}_{k|k}^{\top} & \hat{\text{v}}_{k|k}^{\top} & \hat{b}_{\omega,k|k}^{\top} & \hat{b}_{\alpha,k|k}^{\top}\end{bmatrix}^{\top}\in\mathbb{R}^{17}
\]
Lastly, the covariance is updated using the following relation 
\begin{equation}
	P_{k|k}=P_{k|k-1}-L_{k}P_{k}^{z}L_{k}^{\top}\in\mathbb{R}^{15\times15}\label{Cov_update}
\end{equation}
\textcolor{black}{The computational complexity of the proposed DQUKF
	is dominated by covariance propagation and matrix operations. For
	an augmented state dimension $L$, the filter requires $2L+1$ sigma
	points, leading to a propagation cost that scales as $\mathcal{O}(L^{2})$.
	However, the dominant computational burden arises from covariance
	updates, SVD-based square root computation, and matrix inversion in
	the Kalman gain, which scale as $\mathcal{O}(L^{3})$. Additionally,
	the measurement update depends on the number of observed features,
	with complexity scaling as $\mathcal{O}(n_{z}^{3})$. Compared to
	EKF-based approaches, the DQUKF incurs higher computational cost due
	to multiple sigma point evaluations, but avoids Jacobian computation
	and provides improved performance in nonlinear estimation.}

\section{VIO framework\label{sec:VIO-framework}}

In most filtering frameworks, the landmarks used in the predicted
measurement model \eqref{pred_measurement} are typically treated
as fixed quantities. This means that the filter's visual update becomes
a direct observation of pose relative to those anchors. This makes
it a map-based Visual Inertial-Navigation (VIN). To transition from
VIN to a VIO framework, the fixed anchors are replaced by visual features
extracted from incoming images. The camera should be employed to correct
the drift imposed on the IMU propagated states. As camera images arrive,
a feature detection algorithm is used to extract image features such
as corners and edges as illustrated in figure \ref{feat_detect}.

\begin{figure*}[!ht]
	\centering{}\includegraphics[width=1\textwidth]{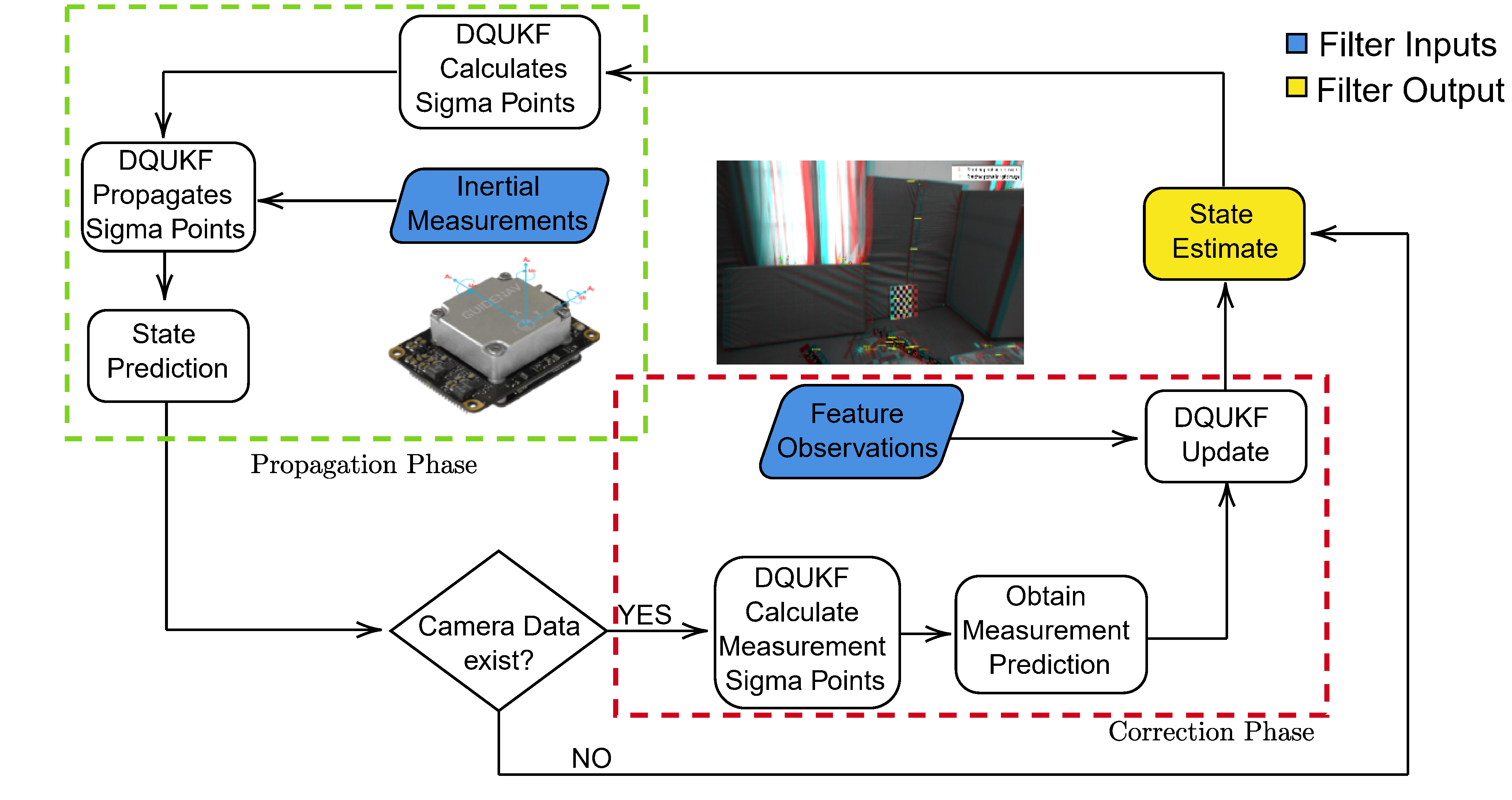} \caption{\label{framework}Schematic of the proposed DQUKF.}
\end{figure*}

Numerous feature detection methods have been proposed in the literature,
including Harris, SIFT, SURF, FAST, and ORB, each offering different
trade offs between computational complexity, invariance properties,
and matching robustness \cite{Rublee2011}. In this work, however,
the focus is placed on a detector that provides stable and well conditioned
features for frame to frame tracking within a stereo video stream.
For a MAV moving within a confined indoor environment, the camera
provides images from a synchronized video stream at fixed scale with
small relative frame to frame motion. Therefore, scale invariance
and descriptor based matching are not primary requirements. Instead,
it is important to obtain stable, repeatable corners that can be reliably
tracked across frames. Since tracking stability is prioritized, the
proposed DQUKF framework employs Shi-Tomasi feature detector, which
can be interpreted as a refinement of the Harris corner detector \cite{Shi1993}.
Rather than relying on the Harris response function, Shi-Tomasi directly
evaluates the eigenvalues of the image structure tensor and selects
points with strong gradients in two orthogonal directions. For a local
image window $W$ the second-moment matrix is defined as 
\[
H=\sum_{(x,y)\in W}\begin{bmatrix}I_{x}^{2} & I_{x}I_{y}\\
	I_{x}I_{y} & I_{y}^{2}
\end{bmatrix}
\]
where $I_{x}$ and $I_{y}$ denote the image intensity gradients along
the horizontal and vertical directions, respectively. Let $\lambda_{1}$
and $\lambda_{2}$ be the eigenvalues of $H$. A pixel location is
classified as a corner if $\min(\lambda_{1},\lambda_{2})>\tau$. where
$\tau$ is a predefined threshold. This criterion ensures that both
eigenvalues are sufficiently large, meaning that the intensity variation
is significant in both directions. After extracting the features,
sparse stereo feature-based matching is performed between the rectified
left and right images. Since the stereo pair is rectified, valid correspondences
are constrained by the epipolar geometry and are expected to lie along
corresponding image rows. For each detected feature in one stereo
image, the corresponding location in the other rectified image is
obtained using point tracking. In this work, point tracking is performed
using KLT optical flow. KLT assumes that small patches of pixels look
very similar in consecutive frames. It tracks features by minimizing
the sum of squared differences in pixel intensity between patches
in frame $k$ and $k+1$. For a small window $W$, KLT solves the
following least squares problem 
\begin{equation}
	E(u,v)=\sum_{(x,y)\in W}\left[I(x+u,y+v,k+1)-I(x,y,k)\right]^{2}
\end{equation}
where $I(x,y,k)$ represents the image intensity at pixel coordinates
$(x,y)$ in frame $k$ and $(u,v)$ represent the horizontal and vertical
displacement of pixels between frame $k$ and $k+1$. Geometric constraints
are then used to filter out poorly tracked features. This is done
by running a RANSAC algorithm \cite{Lee2011}. The remaining valid
feature pairs are tracked across subsequent camera frames using the
same KLT-based point tracking procedure. These valid stereo correspondences
are then used for triangulation. To incorporate the resulting visual
features into the filter update, the measurement model in \eqref{meas_model}
requires the feature positions to be expressed in the world frame.
Given the valid 2D stereo correspondences, the corresponding 3D feature
positions in the camera frame are first obtained through stereo triangulation.
Using the rectified stereo images, the feature depth is computed with
respect to the left camera as 
\begin{equation}
	\mathrm{z}=\frac{f_{x}B}{d}
\end{equation}
where $f_{x}$ is the focal length in the $x$ direction, $B$ is
the stereo baseline distance and $d$ is the disparity. With the depth
value, the 2D pixel coordinates of the features can be projected into
3D space using the following relations. 
\begin{equation}
	\mathrm{x}=\frac{u_{L}-c_{x}}{f_{x}}\mathrm{z}\hspace{3mm}\text{and}\hspace{3mm}\mathrm{y}=\frac{v-c_{y}}{f_{y}}\mathrm{z}
\end{equation}
where ($u_{L}$, $v$) are the 2D feature location and ($c_{x}$,
$c_{y}$) denotes the principal point of the image. It is worth noting
that for a rectified stereo pair, both focal lengths $f_{x}$ and
$f_{y}$ are the same. The resulting 3D feature vector in camera frame
can then be expressed as 
\begin{equation}
	p_{f}^{\mathcal{C}}=\left[\mathrm{x},\mathrm{y},\mathrm{z}\right]^{\top}=\mathrm{z}K^{-1}\left[u_{L},v,1\right]^{\top}
\end{equation}
with $K$ denoting the intrinsic calibration matrix of the left camera.
Finally, the feature position in the world frame is obtained by transforming
from the camera frame using the following relation 
\begin{equation}
	p_{f}^{\mathcal{W}}=R_{\mathcal{C}}^{\mathcal{W}}p_{f}^{\mathcal{C}}+p_{\mathcal{C}}^{\mathcal{W}}\in\mathbb{R}^{3}\label{feat_world_frame}
\end{equation}
where $R_{\mathcal{C}}^{\mathcal{W}}$ is the rotation matrix from
camera to world frame, $p_{f}^{\mathcal{C}}$ is the feature location
in the camera frame and $p_{\mathcal{C}}^{\mathcal{W}}$ is the 3D
coordinates of the camera center in the global frame. Since the camera
and IMU are rigidly attached, the camera center position in the world
frame is taken directly from the propagated IMU states. Performing
the previous transformations to all features yields 
\begin{equation}
	p_{f,i}^{\mathcal{W}}=\ \left[p_{f,1}^{\mathcal{W}},p_{f,2}^{\mathcal{W}},\ldots,p_{f,m}^{\mathcal{W}}\right]\label{feat_matrix}
\end{equation}
where $m$ is the total number of stereo feature matches in the current
image frame. This way, the dimension of \eqref{feat_matrix} becomes
$p_{f,i}^{\mathcal{W}}\in\mathbb{R}^{3\times n_{x}}$, which is similar
to that in \eqref{landmarks}. Using the 3D position of the features
from \eqref{feat_matrix}, the measurement model in \eqref{meas_model}
can be rewritten as 
\begin{equation}
	z_{k}=\text{vec}\left(\mathcal{R}_{q}(q_{k})^{\top}(p_{f,i}^{\mathcal{W}}-p_{k})\right)\in\mathbb{R}^{3n_{x}\times1}
\end{equation}
The filter update step then proceeds as described in Section \eqref{sec:Dual-Quaternion-based-UKF}.
Figure \ref{framework} summarizes the proposed filter with lines
showing how information is being processes inside the DQUKF. \textcolor{black}{The
	framework in figure \ref{framework} follows a centralized estimation
	architecture, which is standard for single-agent visual inertial navigation.
	Distributed fusion strategies are primarily relevant to multi-agent
	systems and are beyond the scope of this work.}

\begin{algorithm}[!h]
	\caption{\label{alg:Alg_QNUKF}DQUKF with VIO for Navigation.}
	{ 
		\textbf{Input}: 
		\begin{enumerate}
			\item[{1:}] UKF parameters $\alpha,\beta,\kappa,\gamma,\lambda$, weights $W_{i}^{(m)},W_{i}^{(c)}$,
			process covariance $Q$, and measurement covariance $R$. 
		\end{enumerate}
		\textbf{Initialization}: 
		\begin{enumerate}
			\item[{2:}] $\hat{x}_{0|0}=[\hat{\tilde{q}}_{0}^{\top}, \hat{\text{v}}_{0}^{\top}, \hat{b}_{\omega_{0}}^{\top}, \hat{b}_{\alpha_{0}}^{\top}]^{\top}$
			, $P_{0|0} \in \mathbb{R}^{15 \times 15}$ (see \eqref{xhat0}, \eqref{P0}) and set $k=1$ 
		\end{enumerate}
		\textbf{while IMU data exists} 
		\begin{enumerate}
			\item[{3:}] $\delta\hat{x}_{k-1|k-1}^{a}=[\delta\hat{x}_{k-1|k-1}^{\top},0_{6\times1}^{\top}]^{\top}$  \textcolor{blue}{\textit{/{*} Prediction {*}/} \textbf{Augmentation}} 
			\item[{4:}] $P_{k-1|k-1}^{a}=\begin{bmatrix}P_{k-1|k-1} & 0_{15\times6}\\
				0_{6\times15} & Q_{n}
			\end{bmatrix}$ (see \eqref{x_aug}, \eqref{p_aug}) 
			\item[] \textbf{Sigma Point Generation} 
			\item[{5:}] $\mathcal{X}_{k-1|k-1}^{a}=\{\delta\hat{x}_{k-1|k-1}^{a},\delta\hat{x}_{k-1|k-1}^{a}\pm\gamma\sqrt{P_{k-1|k-1}^{a}}\}$ 
			\item[{6:}] map error-state sigma points to nominal state using $\mathcal{T}^{-1}$ (see \eqref{dq_pert},\eqref{pose_nom}) 
			\item[{7:}] \textbf{for} each sigma point $i$ 
			\item[] $\qquad\tilde{q}^{(i)}_{k|k-1}=\exp\!\left(\frac{\Delta t}{2}M^{(i)}_k\right)\tilde q^{(i)}_{k-1|k-1}$ 
			\item[] $\qquad\text{v}^{(i)}_{k|k-1}=\text{v}^{(i)}_{k-1|k-1}+(g+\mathcal{R}_q(q^{(i)}_{k-1|k-1})a^{(i)}_k)\Delta t$ 
			\item[] \qquad{}Bias terms remain constant (see \eqref{eq:discrete_dynamics}, \eqref{mat_exp}) 
			\item[] \textbf{end for} 
			\item[{8:}] Compute predicted mean 
			\item[{9:}] Find $\hat{\tilde{q}}_{k|k-1}$ by mapping $\delta \tau^{\text{avg}}_{k|k-1} = \sum_{i=0}^{n-1}W_{i}^{m}\delta \tilde\tau^{(i)}_{k|k-1}$, and $\hat{\text{v}}_{k|k-1}$, $\hat b_{\omega, k|k-1}$, $\hat b_{\alpha, k|k-1}$ using weighted sums (see \eqref{dq_ref}-\eqref{non_dual_avg}) 
			\item[{10:}] Compute predicted covariance $P_{k|k-1}$ 
			(see \eqref{P_pred}, \eqref{sigma-pred})
			\item[] \textbf{if image data available} \textcolor{blue}{\textit{/{*} Update {*}/} \textbf{Measurement Prediction}} 
			\item[{11:}] \qquad{}\textbf{for} each sigma point $i$ 
			\item[] \qquad{}\qquad{}Triangulate stereo features to obtain $p_{f}^{\mathcal{C}}$ 
			\item[] \qquad{}$\qquad p_{f}^{\mathcal{W}}=R_{\mathcal{C}}^{\mathcal{W}}p_{f}^{\mathcal{C}}+p_{\mathcal{C}}^{\mathcal{W}}$ 
			\item[] \qquad{}$\qquad\mathcal{Z}^{(i)}_{k}=\text{vec}\left(\mathcal{R}_q(\mathcal{X}_{k|k-1}^{a,q^{(i)}})^{\top}(p_{f,i}^{\mathcal{W}}-\mathcal{X}_{k|k-1}^{a,\text{t}^{(i)}})\right)$ 
			\item[] \qquad{}\textbf{end for} 
			\item[{12:}] \qquad{}$\hat{z}_{k}=\sum W_{i}^{(m)}\mathcal{Z}^{(i)}_{k}$ 
			\item[{13:}] \qquad{}$P_{k}^{z}=\sum W_{i}^{(c)}(\mathcal{Z}^{(i)}_{k}-\hat{z}_{k})(\mathcal{Z}^{(i)}_{k}-\hat{z}_{k})^{\top}+R$ 
			\item[{14:}] \qquad{}$P_{k|k-1}^{xz}=\sum W_{i}^{(c)}(\mathcal{X}_{k|k-1}^{a}-\hat{x}_{k|k-1})(\mathcal{Z}^{(i)}_{k}-\hat{z}_{k})^{\top}$ 
			\item[] \qquad{}\textbf{Update Step} 
			\item[{15:}] \qquad{}$L_{k}=P_{k|k-1}^{xz}(P_{k}^{z})^{-1}$ 
			\item[{16:}] \qquad{}$\psi_k=L_{k}(z_{k}-\hat{z}_{k})$ 
			\item[{17:}] \qquad{}$\delta \tilde q_k = \mathcal{T}^{-1}(\psi^{\tau}_k)\rightarrow \hat{\tilde q}_{k|k} = \hat{\tilde q}_{k|k-1} \otimes \delta \tilde q_k$  
			\item[{18:}] \qquad{}$\hat{x}_{k|k}=\hat{x}_{k|k-1}+L_{k}(z_{k}-\hat{z}_{k})$ 
			(see \eqref{upd_dq}, \eqref{upd_non_dq}) 
			\item[{19:}] \qquad{}$P_{k|k}=P_{k|k-1}-L_{k}P_{k}^{z}L_{k}^{\top}$ 
			\item[] \textbf{else} 
			\item[] \qquad{}\textbf{No visual update} 
			\item[{20:}] \qquad{}$\hat{x}_{k|k}=\hat{x}_{k|k-1}$ \qquad{} and \qquad{}$P_{k|k}=P_{k|k-1}$ 
			\item[] \textbf{end if} 
			\item[{24:}] $k=k+1$ 
		\end{enumerate}
		\textbf{end while}
	}
\end{algorithm}

\section{Validation\label{sec:Results}}

\begin{figure}[!ht]
	\centering{}\includegraphics[width=1\linewidth]{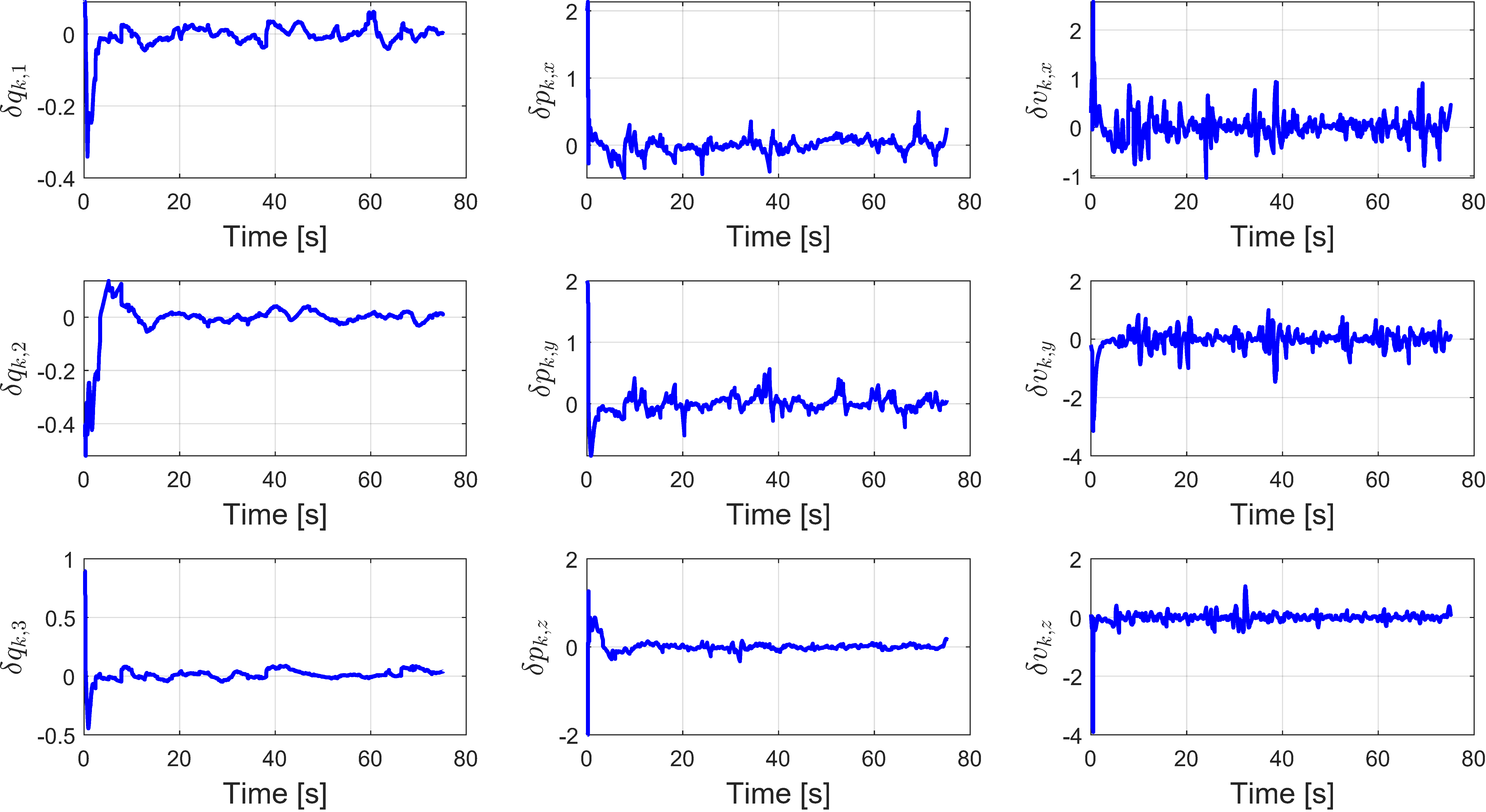} \caption{\label{errors}MAV navigation error}
\end{figure}

\begin{figure*}[!ht]
	\centering{}\includegraphics[width=1\textwidth]{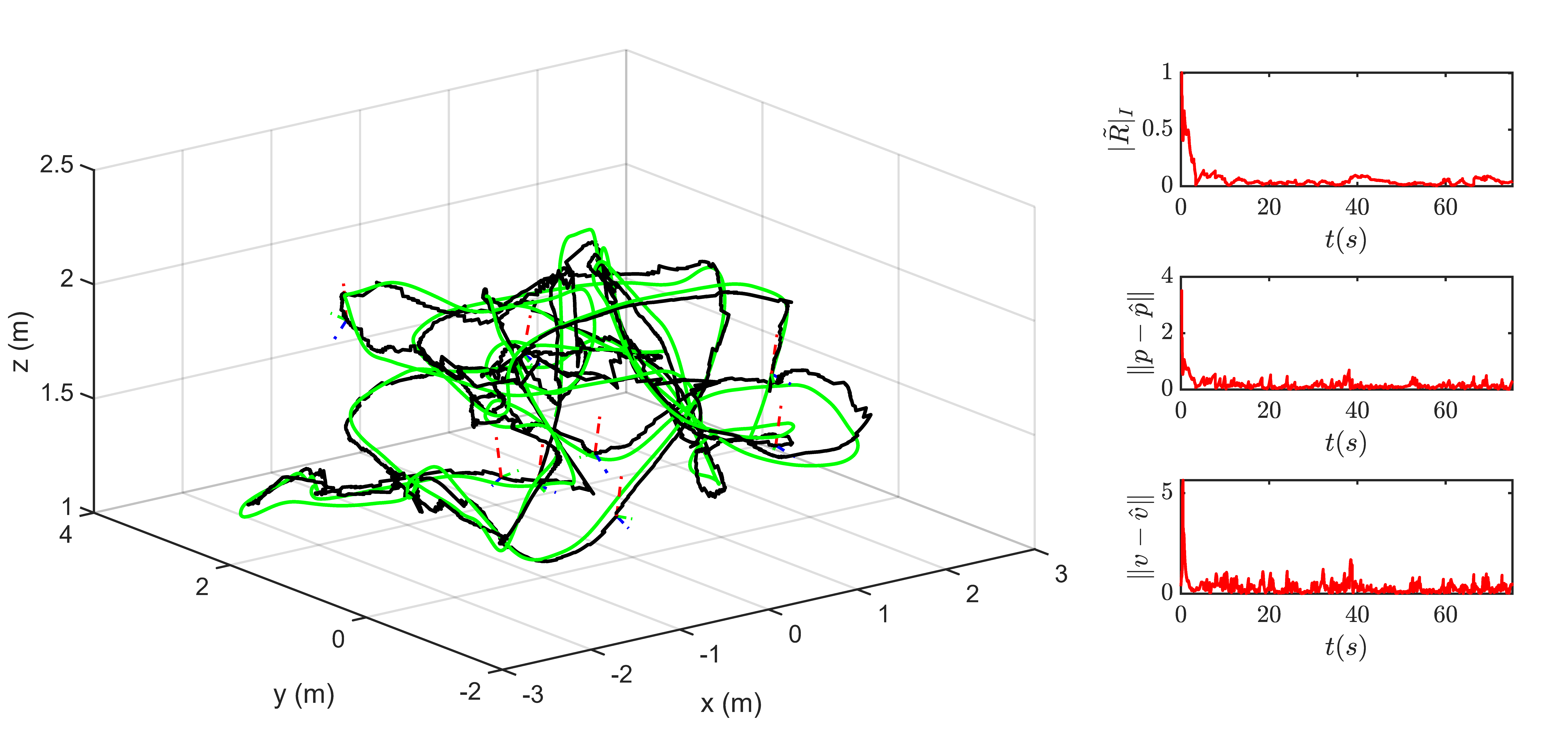}
	\caption{\label{trajectory}3D navigation trajectory of the MAV for experiment 2 (medium). The estimated trajectory (black) is overlaid with the ground truth (green). The corresponding attitude, position, and velocity error magnitudes are presented on the right.}
\end{figure*}

\begin{figure}[H]
	\centering{}\includegraphics[width=1\linewidth]{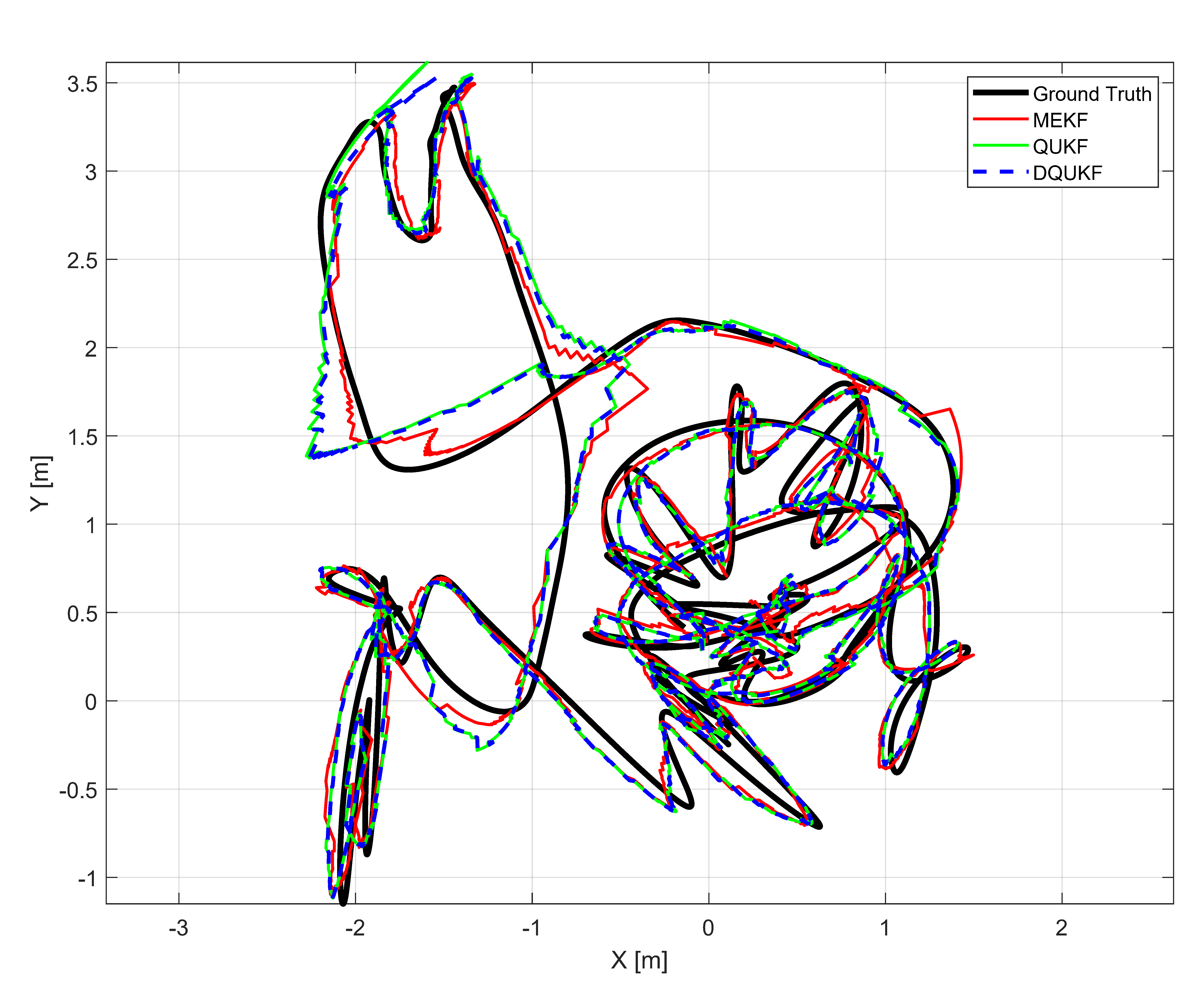} \caption{\label{top_down} Top-down (XY-plane) MAV trajectory comparison with ground truth.}
\end{figure}

\begin{figure}[!ht]
	\centering{}\includegraphics[width=1\linewidth]{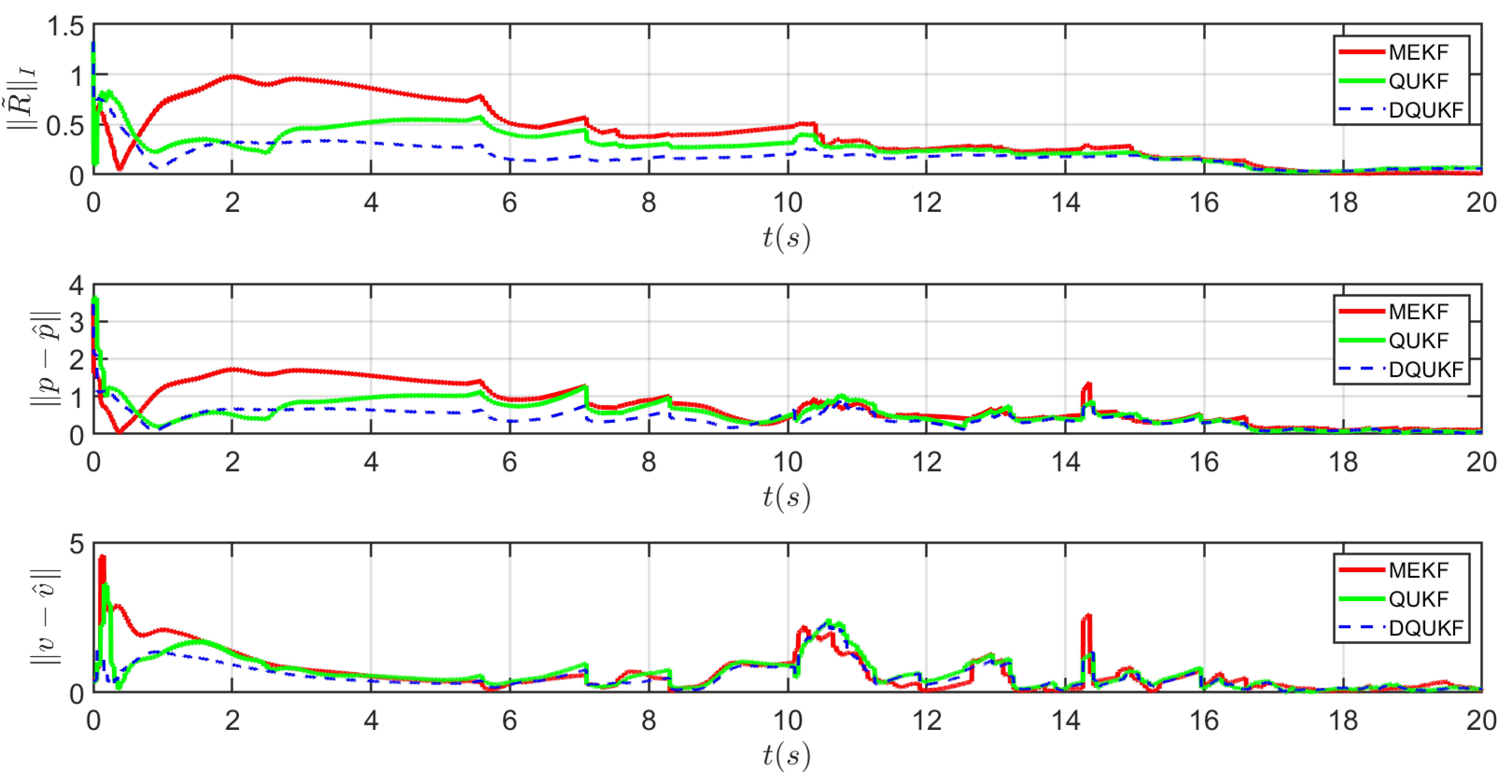} \caption{\label{Transient}Error magnitude comparison between
		MEKF, QUKF and DQUKF (proposed) For Experiment 3.}
\end{figure}

To assess the performance of the proposed DQUKF navigation algorithm,
experiments were conducted using the EuRoC dataset \cite{Burri2016}.
This dataset contains recordings from an Ascetec Firefly hex-rotor
MAV operating in a static indoor environment, including IMU measurements,
stereo images, and ground truth data. The stereo images consist of
simultaneous monochrome frames captured by an Aptina MT9V034 global
shutter sensor at 20 Hz. The MAV's linear acceleration and angular
velocity were measured by an ADIS16448 IMU at 200 Hz. Since the camera
and IMU operate at different sampling frequencies, image measurements
are not available at every IMU timestamp. To handle this, the DQUKF
navigation algorithm performs the update step only when image data
is available; otherwise, the estimated state is propagated, setting
$\hat{x}_{k|k}$ equal to the predicted state $\hat{x}_{k|k-1}$ as
shown in figure \ref{framework}. To assess the filter performance,
the estimated state vector must closely match the ground truth data
provided in the dataset. The navigation error can thus be defined
component wise as 
\begin{equation}
	\begin{cases}
		\tilde{R}_{k} & =q_{k}\ominus\hat{q}_{k}\\
		p_{e,k} & =p_{k}-\hat{p}_{k}\\
		\text{v}_{e,k} & =\text{v}_{k}-\hat{\text{v}}_{k}
	\end{cases}\label{error}
\end{equation}
These errors provide detailed insight into the estimation performance
of each state variable, allowing axis specific analysis of attitude,
position, and velocity. To provide a benchmark for future studies,
the proposed filter is tested on three different datasets: V1\_01\_easy,
V1\_02\_medium, V1\_03\_difficult. Figure \ref{errors} shows the
specific axis errors for orientation (column 1), position (column
2) and velocity (column 3) for the V1\_02\_medium sequence. From the
figure it is clear that all errors converge to near zero, which indicates
that the estimated quantities are in track with the ground truth.
In addition to these individual errors, an overall scalar performance
metric can be defined through the error magnitude. This magnitude
compresses the error information (e.g. $p_{x},p_{y},p_{z}$) into
a single quantity, enabling straightforward comparison of filter accuracy
and convergence behavior across different estimators or datasets.
Figure \ref{trajectory} illustrates the filter estimated navigation
trajectory for the MAV inside the GPS denied confined space of a room. 
The navigation error magnitudes, i.e. $||\tilde{R}_{k}||,||p_{e,k}||,||\text{v}_{e,k}||$,
on the right of the trajectory plot highlight the effectiveness of
the proposed DQUKF. Although the initial estimation error is significant,
the filter stabilizes quickly, achieving near-zero error within a few seconds. 
\textcolor{black}{Another important quantity that is useful in assessing the filter's
	performance is the Root Mean Squared Error (RMSE). The RMSE can be defined as 
	\begin{equation}
		\text{RMSE}=\sqrt{\frac{1}{m_{k}}\sum_{k=1}^{m_{k}}e_{k}^{2}}\label{RMSE}
	\end{equation}
	where, $e_k$ is the magnitude of the errors in \eqref{error}. Notice that the biases were not included in RMSE calculation as the primary focus is the navigation problem, which is sufficiently
	described by the orientation, position and velocity.}
\textcolor{black}{Table~\ref{tab:Performance_comparison} shows the RMSE comparison between the Multiplicative EKF (MEKF), Quaternion UKF (QUKF) and the DQUKF. Experiments 1, 2 and 3 refer to the
	easy, medium and difficult datasets respectively.} 
\begin{table}[h!]
	\centering
	\caption{Performance comparison of DQUKF (Proposed) vs MEKF and QUKF on EuRoC datasets.}
	\label{tab:Performance_comparison}
	
	\setlength{\tabcolsep}{3pt} 
	\renewcommand{\arraystretch}{0.9}
	
	\resizebox{\columnwidth}{!}{
		\begin{tabular}{lccccccccc}
			\toprule
			& \multicolumn{3}{c}{Experiment 1} & \multicolumn{3}{c}{Experiment 2} & \multicolumn{3}{c}{Experiment 3} \\
			\cmidrule(lr){2-4} \cmidrule(lr){5-7} \cmidrule(lr){8-10}
			Filter 
			& Att. & Pos. & Vel. 
			& Att. & Pos. & Vel. 
			& Att. & Pos. & Vel. \\
			\midrule
			MEKF     
			& 0.5742 & 2.0246 & 2.4397 
			& 0.1484 & 0.3346 & 0.6625 
			& 0.2331 & 0.4340 & 0.4769 \\[6pt]
			
			QUKF     
			& 0.2676 & 1.6258 & 1.6084 
			& 0.0806 & 0.2764 & 0.4459 
			& 0.1519 & 0.3347 & 0.4748 \\[6pt]
			
			DQUKF   
			& 0.2893 & 1.0280 & 1.0261 
			& 0.1112 & 0.2843 & 0.5011 
			& 0.1053 & 0.2584 & 0.4237 \\
			\bottomrule
		\end{tabular}
	}
\end{table}
\textcolor{black}{From Table~\ref{tab:Performance_comparison}, the
	baseline filters (MEKF and QUKF) exhibit a broader spread in RMSE
	values compared to the proposed DQUKF, particularly in the more challenging
	scenarios. In Experiment 1, the baselines show large errors, with
	attitude ranging from approximately 0.27 to 0.57 rad, position from
	1.63 to 2.02 m, and velocity from 1.61 to 2.44 m/s. The DQUKF achieves
	the lowest RMSE in both position and velocity, with a substantial
	reduction compared to the baselines, while maintaining competitive
	attitude accuracy close to the QUKF. This indicates a clear advantage
	of the proposed method under aggressive initialization conditions.
	In Experiment 2, all filters operate within a tighter performance
	range, with attitude below 0.15 rad, position below 0.34 m, and velocity
	below 0.66 m/s. The QUKF achieves the lowest RMSE across all three
	states, while the DQUKF remains competitive but does not consistently
	outperform the baselines in this scenario. In Experiment 3, the spread
	increases again, with attitude ranging from 0.10 to 0.23 rad, position
	from 0.26 to 0.43 m, and velocity from 0.42 to 0.48 m/s. The DQUKF
	achieves the lowest RMSE in all three categories, outperforming both
	MEKF and QUKF, indicating improved robustness under more challenging
	conditions. Overall, the DQUKF exhibits more consistent performance
	across experiments, with clear gains in position and velocity estimation
	while maintaining competitive attitude accuracy. The improvement is
	most evident under aggressive initialization, where the increased
	nonlinearity challenges the MEKF and QUKF. In these cases, all filters
	exhibit strong transient oscillations before convergence, which elevates
	the reported RMSE. With less aggressive initialization, the transient
	response becomes shorter and less pronounced since the filters start
	closer to the true state and require less correction. Thus, the RMSE
	values reported in the table would be lower. To complement the RMSE
	results, a top-down (XY-plane) trajectory comparison in figure \ref{top_down}
	is provided to visually illustrate horizontal drift and path consistency
	of the MEKF, QUKF, and proposed DQUKF relative to the ground truth
	for V1\_03\_difficult sequence.} 
\noindent \textcolor{black}{figure \ref{Transient} illustrates the temporal evolution in the transient phase of the attitude error magnitude (top), position error magnitude (middle), and velocity error magnitude
	(bottom) for the V1\_03\_difficult sequence, comparing the MEKF, QUKF,
	and the proposed DQUKF.}
\textcolor{black}{As observed, the MEKF exhibits the largest transient
	errors, especially in the attitude and position components, and requires
	a longer interval to settle compared with the unscented filters. The
	QUKF improves the initial response relative to the MEKF, but still
	shows noticeable fluctuations and error peaks during the transient
	phase. In contrast, the proposed DQUKF maintains lower attitude and
	position errors over most of the transient interval and demonstrates
	a smoother convergence trend. The velocity errors of the three filters
	become more comparable after the initial transient, although the DQUKF
	generally avoids some of the larger spikes observed in the baseline
	filters. At steady state, all three filters eventually converge to
	near-zero error with only small residual fluctuations. To evaluate
	the computational complexity, Table~\ref{tab:computational_complexity}
	reports the total execution time, the average time per iteration,
	and the relative computational cost for experiment 3.} 
\begin{table}[h!]
	\centering
	\caption{Average Computational Cost Of The Evaluated Filters.}
	\label{tab:computational_complexity}
	\setlength{\tabcolsep}{5pt}
	\renewcommand{\arraystretch}{1.1}
	\begin{tabular}{lccc}
		\toprule
		Filter & Total Time (s) & Avg. Time/Step (ms) & Relative Cost \\
		\midrule
		MEKF  & 95  & 5.04  & 1.00$\times$ \\
		QUKF  & 330 & 17.52 & 3.47$\times$ \\
		DQUKF & 180 & 9.55  & 1.89$\times$ \\
		\bottomrule
	\end{tabular}
\end{table}
\textcolor{black}{The MEKF is the most computationally efficient among
	the evaluated filters. The DQUKF requires more computation than the
	MEKF, but remains less computationally demanding than the QUKF for
	the considered simulation. Lastly, to highlight the robustness of
	the established VIO framework, the attitude, position, and velocity
	RMSE are computed under varying feature availability, as shown in
	Table~\ref{tab:feature_degradation}. Relative to the 60-feature
	baseline, the 40-feature case shows only a modest change in performance,
	with slightly higher attitude and position RMSE but lower velocity
	RMSE. A clearer degradation appears at 20 features, where the attitude
	and position RMSE increase to 0.1829 rad and 0.3507 m, respectively,
	indicating that estimation accuracy deteriorates when the number of
	reliable visual constraints becomes limited. The RMSE in the 10-feature
	case does not increase further because the feature set is repeatedly
	reinitialized once the number of tracked features falls below six.
	This frequent reinitialization limits long-term error accumulation,
	so the extreme case should not be interpreted as improved observability.
	Overall, the results show that the DQUKF remains stable under reduced
	feature availability, while performance degradation becomes evident
	as visual measurement constraints are depleted.} 
\begin{table}[h]
	\centering
	\setlength{\tabcolsep}{10pt}
	\caption{DQUKF Estimation performance under feature degradation}
	\label{tab:feature_degradation}
	\begin{tabular}{lcccc}
		\hline
		\textbf{Case} & \textbf{Features} & \textbf{Attitude} & \textbf{Position} & \textbf{Velocity} \\
		\hline
		Baseline & 60 & 0.1053 & 0.2584 & 0.4237 \\
		Moderate & 40 & 0.1229 & 0.2699 & 0.3672 \\
		Severe & 20 & 0.1829 & 0.3507 & 0.3043 \\
		Extreme & 10 & 0.1242 & 0.3433 & 0.3444 \\
		\hline
	\end{tabular}
\end{table}

\section{Conclusion\label{sec:Conclusion}}

This paper introduced a Dual Quaternion Unscented Kalman Filter integrated
with a tightly coupled Visual Inertial Odometry framework for navigation
in GPS-denied environments. The proposed formulation represents the
nominal pose using unit dual quaternions while describing the local
pose uncertainty through a six-dimensional twistor parameterization.
This provides a geometrically consistent error-state formulation that
avoids applying Euclidean corrections directly to the constrained
dual quaternion pose. By combining high-rate IMU propagation with
visual feature-based updates, the framework corrects inertial drift
and maintains consistent estimation of attitude, position, velocity,
and bias states. Simulation results on the EuRoC MAV dataset showed
that the proposed DQUKF converges under high initialization uncertainty
and maintains reliable accuracy across the tested flight sequences.
The DQUKF achieved the lowest position and velocity RMSE in Experiment
1, with 1.0280 m and 1.0261 m/s, respectively, and achieved the lowest
attitude, position, and velocity RMSE in the difficult sequence, with
0.1053 rad, 0.2584 m, and 0.4237 m/s, respectively. The trajectory
and error plots showed close agreement with the ground truth after
convergence, while the computational and feature-degradation studies
demonstrated moderate computational cost and stable performance with
position RMSE below 0.36 m under reduced visual feature availability.

The results indicate that the proposed framework provides a consistent
and robust filtering structure for visual-inertial navigation in GPS-denied
environments. Future work will focus on (i) developing methods for
adaptively tuning the noise covariance matrices, (ii) incorporating
loop-closure detection to enforce global map consistency, moving the
system toward a full SLAM formulation rather than pure navigation,
and (iii) validate the framework in real-world flight experiments.

\balance
\bibliographystyle{IEEEtran}
\bibliography{DQUKF}

\end{document}